%% file: main.tex
\documentclass[letterpaper]{article} 
\usepackage{aaai25}  
\usepackage{times}  
\usepackage{helvet}  
\usepackage{courier}  
\usepackage[hyphens]{url}  
\usepackage{graphicx} 
\usepackage{natbib}  
\usepackage{caption} 
\usepackage{algorithm}
\usepackage{algorithmic}
\usepackage{amsmath}
\usepackage{todonotes}
\RequirePackage[inline,shortlabels]{enumitem}
\usepackage{newfloat}
\usepackage{listings}
\DeclareCaptionStyle{ruled}{labelfont=normalfont,labelsep=colon,strut=off} 
\floatstyle{ruled}
\newfloat{listing}{tb}{lst}{}
\floatname{listing}{Listing}
\usepackage{xspace}

\title{Synthetic Tabular Data Generation for Imbalanced Classification: The Surprising Effectiveness of an Overlap Class}

\author {
    Annie D'souza\equalcontrib,     
    Swetha M\equalcontrib,     
    Sunita Sarawagi
}
\affiliations {
    Indian Institute of Technology, Bombay
}

\usepackage{bibentry}

\newcommand{\sysname}{ORD\xspace}
\newcommand{\condcol}{OR\xspace}
\newcommand{\CM}{D_{0}\xspace}
\newcommand{\CMM}{D_{00}\xspace}
\newcommand{\CmM}{D_{01}\xspace}
\newcommand{\Cm}{D_{1}\xspace}
\newcommand{\xhdr}[1]{\vspace{1mm}\noindent{{\bf #1.}}}

\begin{document}
\maketitle

\begin{abstract}
Handling imbalance in class distribution when building a classifier over tabular data has been a problem of long-standing interest.  One popular approach is augmenting the training dataset with synthetically generated data.  While classical augmentation techniques were limited to linear interpolation of existing minority class examples, recently higher capacity deep generative models are providing greater promise. 

However, handling of imbalance in class distribution when building a deep generative model is also a challenging problem, that has not been studied as extensively as imbalanced classifier model training.  We show that state-of-the-art deep generative models yield significantly lower-quality minority examples than majority examples.
We propose a novel technique of converting the binary class labels to ternary class labels by introducing a class for the region where minority and majority distributions overlap.  We show that just this pre-processing of the training set, significantly improves the quality of data generated spanning several state-of-the-art diffusion and GAN-based models. While training the classifier using synthetic data, we remove the overlap class from the training data and justify the reasons behind the enhanced accuracy. We perform extensive experiments on four real-life datasets, five different classifiers, and five generative models demonstrating that our method enhances not only the synthesizer performance of state-of-the-art models but also the classifier performance. 
\footnotetext{ 20d070028@iitb.ac.in, 23m0756@iitb.ac.in, sunita@iitb.ac.in}





\end{abstract}

%
\begin{links}
    \link{Code}{https://github.com/Annie2603/ORD}
\end{links}

\section{Introduction}
The problem of imbalance in class distributions is encountered in several fields, including fraud detection in the banking sector, disease diagnosis in the medical sector, and anomaly detection in finance, cybersecurity, and manufacturing industries. Imbalanced data refers to the distribution of classes or categories being highly skewed or disproportionate in a dataset, i.e., one class or category is significantly more prevalent than the others. The presence of imbalance introduces several challenges in training classification models as they are unable to learn the distribution of the rare class. This has been a problem of long-standing interest and many different techniques have been proposed ranging from cost-sensitive loss functions~\cite{costsensitive}, oversampling minority instances and/or under-sampling majority instances~\cite{li2021hybrid}, and augmentation with synthetic instances~\cite{dataGenMethods}.  In this paper, we focus on data augmentation-based techniques since these are particularly effective~\cite{Chawla_2002}. One popular classical data augmentation method, SMOTE, generates new minority instances using a convex combination of existing minority instances.  However, recent deep data generators yield even higher accuracy due to the enhanced quality of generated examples.

Our focus in this paper is tabular data, and recently many high-quality generators have been proposed for generating tabular data ranging from GAN-based models like CTabGAN \cite{ctabgan} to diffusion-based models like TabSyn \cite{zhang2024mixedtype} and ForestFlow \cite{forestflow}
Most of these have been evaluated after training on balanced datasets, where their generated data have been shown to be highly effective in training classifiers that match the accuracy of real data.  We are not aware of any study that evaluates them in highly imbalanced input data.

In this paper, we show that default training of generative models yields significantly worse quality minority instances than majority instances as measured by the label provided by an Oracle Bayesian model. Simple fixes to the problem by over and under-sampling initial real data distributions are not effective.  



We present a method called \sysname\footnote{for Overlapped Region Detection} that employs three key ideas to improve the training of imbalanced classifiers on tabular data using synthetic data generators: (1) We pre-process the available real data by identifying a small subset of the majority instances called the {\em overlap} set that lies on the boundary of majority and minority examples. (2)  We modify the generator to be class conditional and generate three-way labeled instances corresponding to minority, overlapping majority, and clear majority.  (3) We train the final classifier with a careful mix of real and synthetically generated instances comprising of balanced sample of synthetic data generated consisting of only the minority and clear majority classes, and real minority.  

Our method is orthogonal to the type of data generator and classifier, and we evaluate it on four different tabular data generator architectures and five different classification models. We perform a set of insightful ablation studies to tease apart the reason why our method outperforms existing methods.  We conclude that our method yields improved accuracy both due to the generator providing higher quality synthetic minority examples, and the classifier improving due to selective under-sampling of overlapping majority examples.

 

\begin{figure*}
     \centering
     \includegraphics[width=1\linewidth]
     {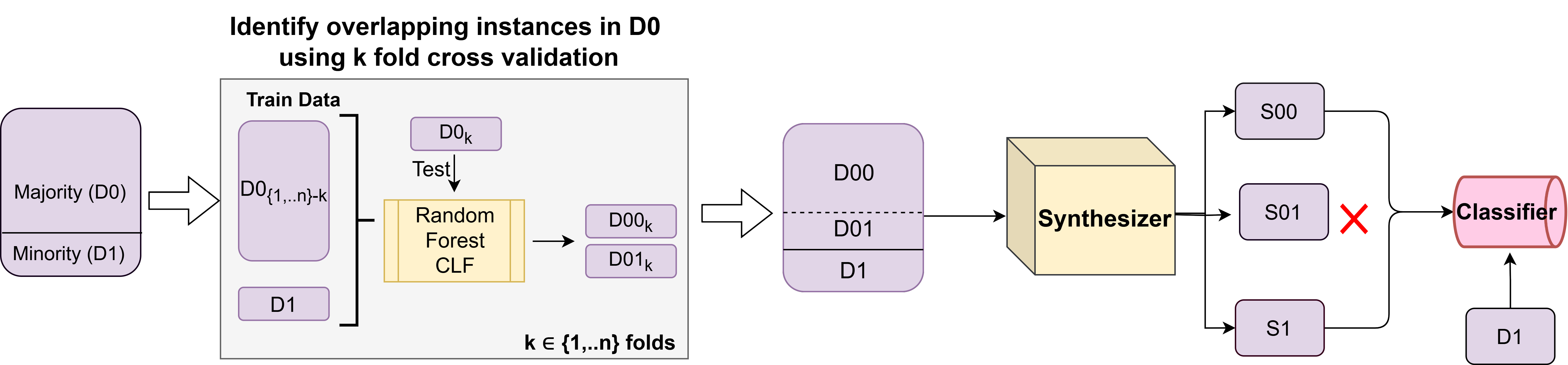}
     \caption{The method of \sysname and its application to generate synthetic data. Synthetic data is then used to train a better classifier compared to when no \sysname is used. Steps in the overall algorithm: 1. To use k-fold training to identify the overlap in the validation set. This uncertainty is labelled as a third class label $\CmM$. 2. Generate synthetic data using the three class dataset instead of binary class. The synthetic data quality is much better as the distribution is better reproduced by Generative models. 3. Train the final classifier with equal proportions of majority and minority synthetic data while discarding the overlapped region $\CmM$ which makes learning the decision boundary easier.}
     \label{fig:method}
 \end{figure*}
 

The main contributions of this paper are as follows:
\begin{itemize}
    \item We evaluate state-of-the-art deep generative models for tabular data on imbalanced training data, and show that the quality of minority examples they generate is much worse than majority examples. 
    \item We propose a new method called \textbf{\sysname} that introduces a new class comprising of the subset of majority examples that lie in the boundary of the two classes, harnesses these to impose a ternary classification loss when training the generative model, and finally trains the classifier using generated examples that lie outside the overlap region.   
    \item We propose a new SOTA model
    \textbf{CTabSyn}, a variation of TabSyn \cite{zhang2024mixedtype} with the inclusion of \sysname and conditional training and generation.
    \item We provide extensive experimental evaluation and proof that \sysname provides 
    significant increase in accuracy.
    \item We justify the enhanced accuracy by showing that our method both improves the quality of minority examples generated from our method, and the classifier gains from selective under-sampling of points in the overlap region.
\end{itemize}

\section{Related Works}
Traditionally, for training classifiers on imbalanced data, either the minority class is oversampled \cite{Chawla_2002} or majority class is undersampled. While efficient to some extent, majority under-sampling methods can discard potentially useful instances.  Under sampling methods include cluster-based \cite{zhang2010cluster} under-sampling and Neighbourhood \cite{vuttipittayamongkol2020neighbourhood} based boundary undersampling.  While over sampling methods include
Tomek link based oversampling~\cite{OBMI}.
Over-sampling can be ineffective when number of minority points is very small.  In such cases, a more effective alternative is generating synthetic examples.  A pioneering method in this category is SMOTE (for
 Synthetic Minority Oversampling Technique)~\cite{Chawla_2002} that interpolates between minority points to generate new minorities. Several variations of SMOTE like BorderlineSMOTE ~\cite{han2005borderline} and ADASYN ~\cite{he2008adasyn}, SMOTE Tomek \cite{SMOTEtomek} also exist.

Recently, deep generative models have provided much higher quality synthetic data across a variety of domains but we specifically focus on synthetic generators for tabular data.
 Conditional Tabular GAN (CTGAN) ~\cite{ctgan} is a Generative Adversarial Network based synthetic data generator.  \cite{HSCGS} proposed to use density-based clustering to filter out noisy and boundary samples before feeding to CTGAN synthesizer. \cite{CTGANENN} proposes to oversample minority with CTGAN and combine with existing majority points. 
 CTABGAN~\cite{ctabgan} and CTABGAN+~\cite{zhao2022ctabgan+} are improvements over CTGAN. They perform training by sampling based on frequency of discrete columns to handle imbalances. Among the transformer-based \cite{transformer} synthesizers, Generation of Realistic Tabular data i.e. GReaT  \cite{GREAT}, TabuLA model~\cite{tabula} and Tabular Masked transformers ~\cite{gulati2023tabmt}, TabLLM \cite{tabllm} are noteworthy synthetic data generators. Normalizing flows are also used to estimate the data distribution \cite{MAF}.

Of late, diffusion-based models are providing higher quality generations of synthetic data.  These include
the Denoising diffusion probabilistic model TabDDPM~\cite{tabddpm}, TabSyn~\cite{zhang2024mixedtype} a Latent diffusion model, and Forestflow~\cite{forestflow} based on XGBoost~\cite{chen2016xgboost}.  We experiment with several categories of existing generators for tabular data, and found  ForestFlow and TabSyn to provide the best accuracy.  

Most existing data synthesizers have not considered \textbf{imbalanced} data. Ours is the first work that harnesses these models for generating synthetic data in the imbalanced class setting.  We show that even the training of the generators is affected by the data imbalance.  Our method \sysname can be applied to all of the existing synthetic generators to enhance its quality of imbalanced data.  


\newcommand{\dsyn}{D_S}
\section{Problem Statement}
Let \textbf{$X=X_1\times\ldots \times X_d$} denote the space of $d$-dimensional record data where each \textbf{$X_j$} could be continuous or categorical.  Let $Y=\{0,1\}$ denote a binary class label space and $P(X,Y)$ denote their unknown joint distribution.  We are provided a labeled dataset $D=\{(x_1,y_1)\ldots (x_N,y_N)\}$ from the distribution where the fraction of instances in $D$ with label 1 (minority class) is significantly smaller than (order of 2\%) $N$.  Our goal is to train a classifier $M:X\mapsto Y$ that provides high accuracy separately for both the minority and majority class.  Instead of depending on the given imbalanced data $D$ alone, we seek to sample synthetic data $\dsyn$ from a generative model $G$ trained on $D$. The classifier may be trained on a mix of synthetic and real data, but it is always evaluated on real data.  A baseline method is using any of the SOTA tabular data generators to sample a balanced and large number of majority and minority examples as $\dsyn$ to train the classifier $M$.  In the following section, we present our proposed method that improves upon this baseline.

\section{The Overlap Region Detection (\sysname) Method}
One of the major sources of difficulty in training a classifier with highly imbalanced class distributions is that the discriminator fails to find any region where the minority examples are more prevalent than the majority examples.  Our proposed \sysname\ method addresses this challenge by increasing the concentration of minority examples in select regions via data generated from a deep generative model.  We achieve this over three steps as outlined in Figure~\ref{fig:method}: (1) First, we identify a subset $\CmM$ of the majority instances $D_0$ in $D$ that overlaps significantly with the minority examples.  We use a time-tested k-fold cross-validation on a random forest classifier for this step, as we elaborate in Section~\ref{sec:overlap}.  The output of this stage is a split of the majority instances $D_0$ into the overlapping majority $\CmM$ and clear majority $\CMM = D_0- \CmM$.  (2) Second, we train a conditional generator $G$ on the ternary labeled data as elaborated in Section~\ref{sec:gen}. We show that the introduction of a finer-grained class label dataset improves the quality of the generated examples in Table \ref{tab:randomoracle} (3) We train the final classifier $M$ on synthetically generated instances $S_1,S_{00}$ from $G$ that are sampled in equal parts from the minority class and clear majority class, respectively.  We show that excluding the overlap majority instances declutters the decision boundary and improves overall accuracy.   We also found it useful to include only the minority instances $D_1$ in the training pool of $M$.  The inclusion of real majority instances did not help improve the accuracy, possibly because the generator is well-trained on the majority instances.  This is somewhat contrary to the results on balanced dataset where classifiers trained on real data outperform those on synthetic ~\cite{tabddpm}.\\


\begin{algorithm}[tb]
\caption{Synthesis and Classification using \sysname}
\label{alg:algorithm}
\textbf{Input}: Imbalanced real data: majority $\CM$, minority $\Cm$. \\
\textbf{Parameter}: Threshold $\tau$, Folds $k$  \\
\textbf{Output}: Classifier $M$.
\begin{algorithmic}[1] 
\STATE Split $D_0$ into k folds $F_1,\ldots F_k$.
\STATE $\CmM,\CMM=\phi$
\FOR {$i = 1$ to $k$}
    \STATE Train $RF_j$ on $D_0 \setminus F_j$ and $D_1$
    \FOR {$x \in F_i$}
        \STATE confidence\_{maj} = $RF_j$.predict\_prob\_0$(x)$
        \IF {confidence\_{maj} $\leq 1-\tau$}
            \STATE Add $x$ to $\CmM$  with class $2$
        \ELSE
            \STATE Add $x$ to $\CMM$ with class 0
        \ENDIF
    \ENDFOR
\ENDFOR
\STATE Train $G$ on ternary labeled data $\CMM \cup \CmM \cup \Cm$ 
\STATE Sample $S_{00}, S_1$ from $G$ for label 0 and 1.
\STATE Train $M$ using $S_{00} \cup S_{1} \cup D_1$ and return $M$


\end{algorithmic}
\label{alg:overlap_find}
\end{algorithm}


 \subsection{Identifying Overlapping Majority Instances}
 \label{sec:overlap}
In this section, we describe the method to identify the subset $\CmM$ of majority instances $D_0$ that we call the overlapping instances. 
Since neighborhood-based methods are not meaningful for high dimensional data, we use the disagreement of ensemble classifiers like random forest \cite{randomforest} to identify overlap.  We depend on k-fold cross-validation to define splits used to train the random forest and obtain disagreement scores. Since we need the scores only on majority instances, we split only the majority instances $D_0$ into $k$ disjoint sets $F_1,\ldots,F_k$.  For each fold $j$, we train a random forest $RF_j$ on $D_0-F_j$ instances as class 0 and $D_1$ as class 1.  We then apply $RF_j$ on the set-aside fold $F_j$ and obtain the fraction $r_j(x)$ of committee members in the random forest that label each instance $x \in F_j$ as class 0. If there is high disagreement, that is $r_j(x) < 1 - \tau$, where $\tau$ is a given threshold, we identify $x$ as lying in the overlap region and add it to the overlap set $\CmM$. 
Across the $k$ folds, each instance in $D_0$ gets evaluated for its potential of being in the overlap set $\CmM$.  The remaining majority instances are put in $\CMM$ which form the clear majority set.  In Figure~\ref{fig:toy-vis} the top left figure shows an original binary dataset with overlap points extracted.  Note that the overlap points are nicely identified as being on the boundary of the majority and minority regions.\\
We take $\tau$ such that, the number of overlapped majority points = min(number of minority points, r\% of majority points). The r above is generator dependent and we find it by checking the performance on the validation set of a dataset with different r values (we used 3\%, 5\%, 7\%, 9\%) and use the best r value with all datasets for a given synthesizer.

\subsection{Conditional Generators With Ternary Labels}
\label{sec:gen}
\begin{figure}
    \centering
    \includegraphics[width=1\linewidth]{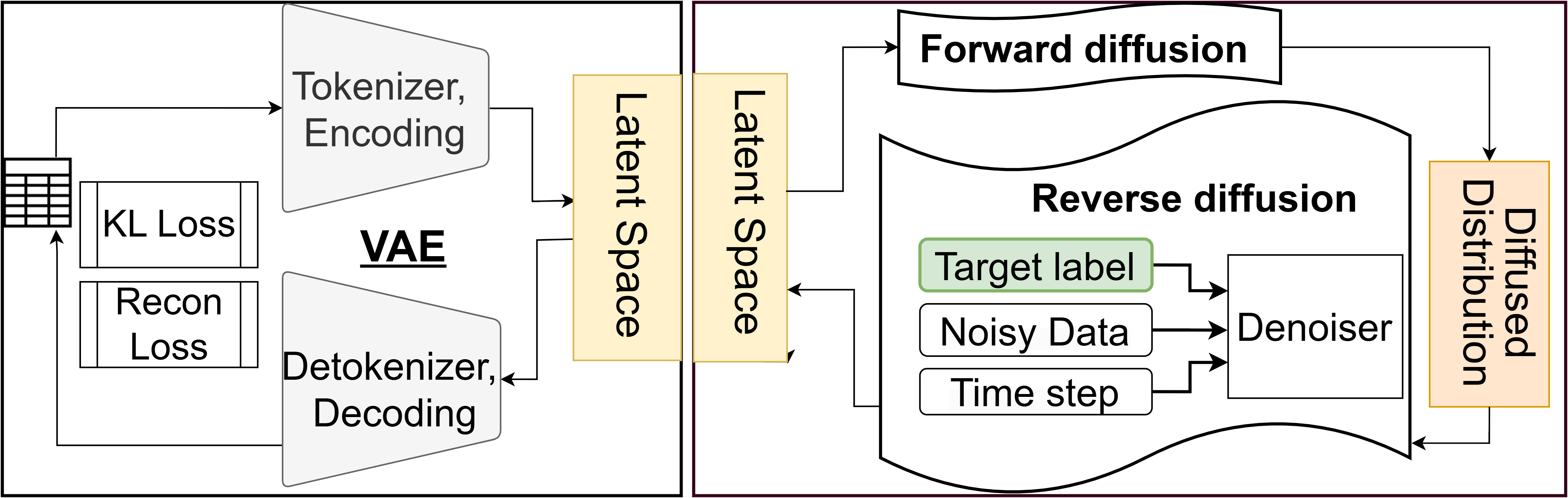}
    \caption{CTabSyn for class conditional tabular data generation.  We needed to make only a small change (highlighted in green)  over existing TabSyn model to improve the quality of generations for imbalanced class distribution and benefit from our finer-grained class labels. Conditional diffusion is implemented by adding the true target embedding as input to denoiser for efficient sampling.}
    \label{fig:ctabsyn}
\end{figure}

Once we assigned finer-grained class labels to the original data $D$, we show in the experiment section that almost all existing deep generative models provided higher quality generators than training an identical model with binary labels.  
The only modification we made was to TabSyn \cite{zhang2024mixedtype} one of the SOTA synthetic tabular data generation models which consists of a diffusion process modeled in a learnt latent space. The tabular data is first passed through a Variational auto-encoder which consists of a tokenizer that generates unique representations for each column which are then fed to a transformer to capture the intricate relations between the columns to obtain data in a continuous latent space instead of separately for numerical and categorical columns. 
This model does not support class conditional generation.  We modified the TabSyn to support conditional generation and call this variant \textbf{CTabSyn}.  The denoiser MLP which takes as input - a noisy version of data ($x_{in}$) and timestep ($t$) of the data now also has an embedding of the true target value ($y$) input to it. 
\begin{equation} \label{eq:process_inputs}
\begin{aligned}
    & t\_emb = \texttt{TimeEmbedding}(t) \\
    & y\_emb = \texttt{Embedding}(y) \\
    & x = \texttt{Linear}(x_{in}) + t\_emb + y\_emb
\end{aligned}
\end{equation}
The overall architecture of CTabSyn is shown in Figure~\ref{fig:ctabsyn}. CTabSyn with \sysname considerably improves the performance of TabSyn as shown in Table \ref{tab:model_performance}.

\section{Experiments} \label{sec:experiments}
We present a detailed evaluation of the performance of \sysname\ vis-a-vis several existing methods of generating synthetic data for training classifiers with imbalanced data.  Our method is agnostic to the type of data generator and classifier, and we perform evaluation on five data generators and five different classifiers. More than the overall results, we dwell on the analysis of the reasons why \sysname\ provides gains.  Here we seek to answer questions like: Are \sysname's gains due to better quality synthetic data generated due to training with finer-grained ternary labels?  Or, is the main reason the selective under-sampling of majority points from the overlapping region? 

\begin{table}[h!]
  \centering
  \begin{tabular}{l|rrrrr}
    \hline
    Dataset & \# Maj & \# Min & \# Unb Min & \%Min/Maj \\ \hline\hline
    Adult   & 34514   & 11208      & 500            & 1.5  \\
    Heloc   & 5559    & 34979      & 700           & 2  \\
    Fintech & 11560   & 11174     & 200           & 1.3  \\
    Cardio  & 35700   & 5000      & 100            & 1.8  \\ 
    Abalone &  3076  &    265   &   265 &     8.6    \\
    Bank &   31970 &     4198  &   4198       & 13.1 \\
    Car &  1324  &    58  &      58     &  4.4\\
    Yeast & 1056   &   127   &     127      & 12 \\ \hline
    
  \end{tabular}
  \caption{Dataset details with count of original majority, original minority and the count of minority in the imbalanced setting with the Minority-Majority ratio for all datasets considered are depicted.}
  \label{tab:dataset_characteristics1}
\end{table}

\paragraph{Datasets}
We evaluate our approach on eight real-world tabular datasets comprising a mix of numerical and categorical features, all with binary target variables. Four datasets—Adult, Heloc, Fintech Users, and Cardio—are originally less imbalanced. To simulate higher imbalance, we undersample the minority class to constitute only 1.5\% to 2\% of the total data. These datasets use a balanced test set for evaluation, while the training set is highly skewed. 

Additionally, we use four inherently imbalanced UCI datasets—Abalone, Bank, Car, and Yeast—without introducing further skew. The test sets for these datasets retain their natural imbalance, reflecting real-world distributions. This combination of balanced and imbalanced test scenarios enables a comprehensive evaluation of the robustness and effectiveness of our method. 
The overall statistics of these datasets and their detailed descriptions including the number of $\CmM$ points for each threshold have been made available in the appendix. 

\begin{figure*}
    \centering
    \includegraphics[width=1\linewidth,height=3.6in]{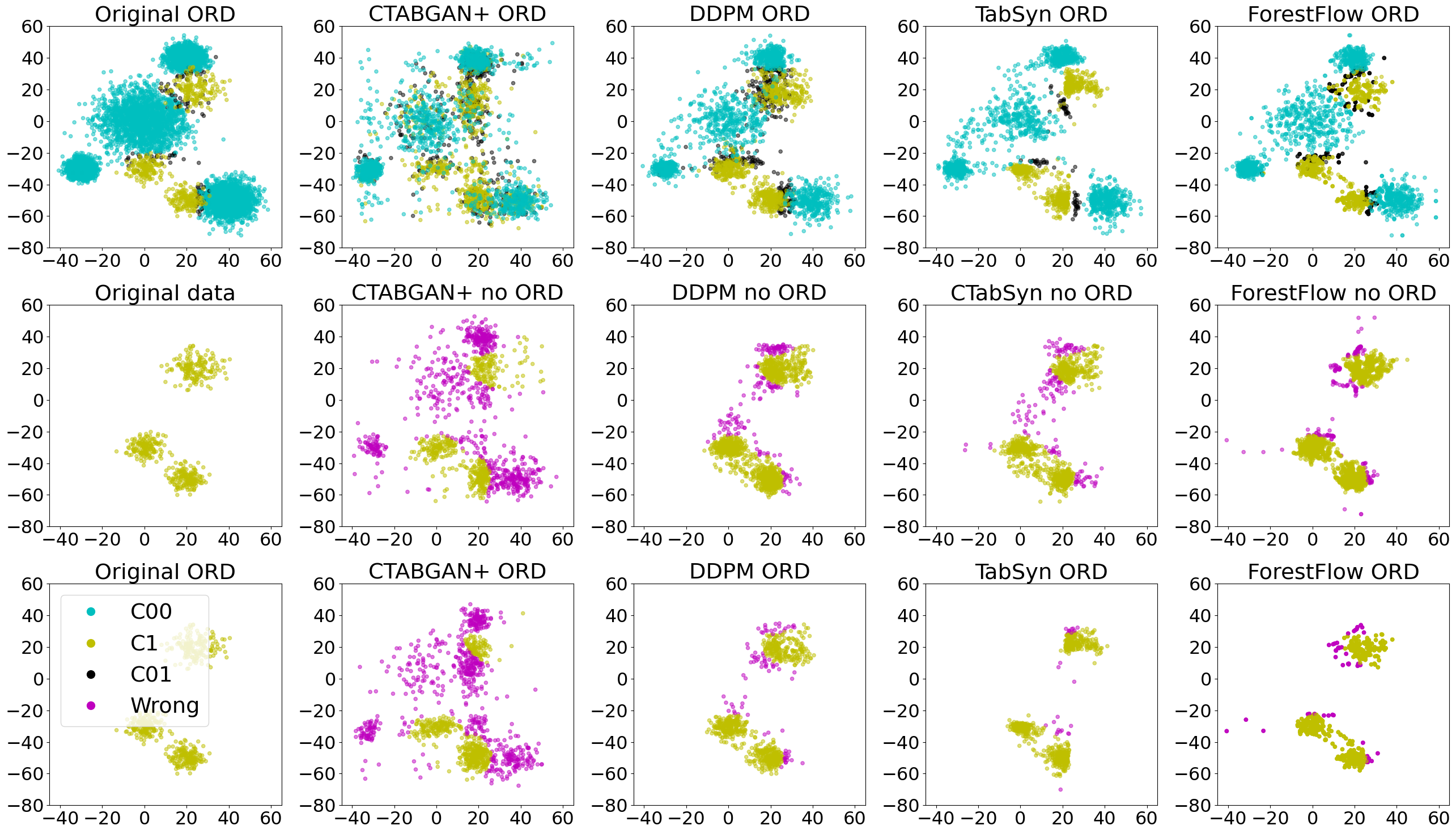}
    \caption{\small Visualisation of Synthetic data for \sysname in 2D datasets. The first row shows data synthesized with \sysname for different Synthesizers along with a clear indication of the overlap class $\CmM$. The second row zooms only on sampled minority with wrong generations marked in pink from generators without \sysname. The third row shows the same data but with \sysname enhanced generators.  The columns correspond to different generators.  Particularly for the diffusion models (last two columns), \sysname\ provides much fewer errors in generated examples than baseline diffusion models, which is already much higher quality than earlier GAN-based generators in columns 2 and 3.}
    \label{fig:toy-vis}
\end{figure*}
\paragraph{Baselines}
Since \sysname is an addendum over synthetic data generation models, we demonstrate the performance improvement it gives over five existing synthetic tabular data generation methods: \textbf{CTGAN, CTABGAN+, TabDDPM, ForestFlow and TabSyn}. In addition, from classical methods, we also include \textbf{SMOTE} and its popular variants i.e. \textbf{borderline SMOTE} \cite{han2005borderline} and \textbf{AdaSyn}. \cite{he2008adasyn}

\paragraph{Evaluation Metrics}
We follow standard practice for evaluating classifiers on imbalanced data and measure accuracy as the macro average of accuracy on \textbf{real} unseen minority and majority instances. This metric is also known as machine learning efficacy. We train five different classifiers:  \textbf{XGBoost, Logistic Regression, Decision Tree, Multi-Layer Perceptron, AdaBoost}.  However, to reduce clutter, we present accuracy grouped as follows: We separately report accuracy on XGBoost since it is generally the best-performing classification method for tabular data, and then we report the average accuracy over the remaining four classifiers. For the test set, we sampled 2000 of each class for the first 4 datasets since they were created by under-sampling from the original datasets. Since the test sets of the UCI datasets are also imbalanced, we additionally report the minority \& majority accuracies, F1 scores and AUCROC in Table \ref{tab:uci_results}

\paragraph{Experimental Setup}
The code for all baseline synthesizers was obtained from their official GitHub repositories. All steps on environment creation, package installation, dataset pre-processing, training, and generation were followed as specified in the repository of the given baseline. Each baseline synthesizer underwent training and generation twice: Once with the original imbalanced dataset to synthesize $\CM$ and $\Cm$ and once with the \sysname processed imbalanced dataset to synthesize $\CMM$, $\CmM$ and $\Cm$.
For each dataset, an equal number of $\Cm$ and $\CM$ synthetic points were used for training the classifier in the case of both with and without \sysname. The reported numbers are an average over 3 runs of synthesizer training. 
We used $\tau$=0.3 for all real datasets and $\tau$=0.2 for the 2D toy datasets. The random forest for overlap detection had 50 trees while the other parameters were used at their default values. $k$=2 was sufficient for the k-fold training.

\begin{table*}[h!]
    \centering
    \begin{tabular}{l|rr|rr|rr|rr}
        \hline
        & \multicolumn{2}{c|}{Adult} & \multicolumn{2}{c|}{Cardio} & \multicolumn{2}{c|}{Fintech} & \multicolumn{2}{c}{Heloc} \\ \cline{2-9}
        Training Data & Avg of 4 & XGBoost & Avg of 4 & XGBoost & Avg of 4 & XGBoost & Avg of 4 & XGBoost \\ \hline\hline
        Unbalanced, original & 47.74 & 60.88 & 32.46 & 50.00 & 33.74 & 66.30 &  30.90 & 58.40 \\
        Smote & 69.34	& 71.45	& 61.62	& 58.77	& 37.90	& 56.00	& 52.27	& 55.90\\
        Borderline Smote & 64.83	& 68.10 & 60.72	& 58.02	& 37.71	& 56.00	& 45.50	& 54.32 \\
        ADASYN & 70.79 & 71.30	& 61.09 & 58.15 & 38.23	& 55.80	& 56.52	& 53.14\\
        TabSyn & 76.87	& 78.28	& 67.38	& 69.56	& \textbf{63.60}	& 63.21	& 65.92	& 65.79  \\
        CTabSyn \sysname & \textbf{80.67}  &	\textbf{81.18} &	\textbf{70.17} &	\textbf{73.10}	& 63.58 &	\textbf{66.46} &	\textbf{66.36}	& \textbf{67.77}  \\ \hline
\hline
        CTGAN & 65.33 & 70.31 & 61.96 & 63.85 & 56.14 & 56.44 & \textbf{62.93}& \textbf{62.86} \\
        CTGAN \sysname & \textbf{72.81} & \textbf{74.52} & \textbf{64.90} & \textbf{66.33} & \textbf{57.58} & \textbf{58.38} & 61.47 & 62.34 \\ \hline
        CTAB-GAN+ & 77.06	& 78.09	& 68.97	& 57.33	& \textbf{57.45}	& 59.40	& 56.26	& 54.77  \\
        CTAB-GAN+ \sysname & \textbf{78.90}	& \textbf{79.72}	& \textbf{69.74}	& \textbf{61.46}	& 54.83	& \textbf{57.66}	& \textbf{63.82}	& \textbf{65.03} \\ \hline
        ForestFlow & 67.79	& 69.64	& 67.42	& 65.25	& 61.19	& 66.22	& \textbf{65.17}	& \textbf{66.80} \\
        ForestFlow \sysname & \textbf{71.20}	& \textbf{72.88}	& \textbf{68.94}	& \textbf{66.03}	& \textbf{62.91}	& \textbf{66.58}	& 64.83	& 66.22\\ \hline
        TabDDPM & 69.55 & 69.76 & 64.91 & 64.02 & 61.88 & \textbf{60.50} & 52.10 & 60.70 \\
        TabDDPM \sysname & \textbf{77.92} & \textbf{78.33} & \textbf{66.87} & \textbf{68.77} & \textbf{63.07} & 57.80 & \textbf{63.22} & \textbf{61.30} \\ \hline
        TabSyn & 76.87	& 78.28	& 67.38	& 69.56	& 63.60	& 63.21	& 65.92	& 65.79  \\
        CTabSyn & 79.80 & 80.01 & 69.45 & 71.80 & \textbf{65.37} & 65.95 & \textbf{66.69} & 65.65 \\
        CTabSyn \sysname & \textbf{80.67} & \textbf{81.18} & \textbf{70.17} & \textbf{73.10} & 63.58 & \textbf{66.46} & 66.36 & \textbf{67.77} \\ \hline

    \end{tabular}
    \caption{Comparison of classification accuracy. First part focuses on existing imbalance handling baselines and our best method CTabSyn \sysname. The next part shows an increase in accuracy for each individual synthesizer using \sysname. Across 26 out of 32 (dataset, generative model) combinations we find that \sysname\ provides a gain in accuracy. }
    \label{tab:model_performance}
\end{table*}

\paragraph{Visualization on 2D Toy Datasets}
We provide a better understanding of the working of our method via visualization of \sysname on a 2D synthetic toy datasets. Detailed information on the creation of these datasets and minority-majority counts has been provided in the appendix.
Dataset Blobs consists of four blobs that belong to the majority $\CM$ and three smaller blobs belonging to the minority $\Cm$. This was created using Scikit Learn Gaussian Blob generation. 

Figure \ref{fig:toy-vis} top left corner shows the original majority instances (cyan) and minority instances (yellow), and the subset of majority that our method marked as Overlap points.  In each of the subsequent plots, we show data generated by \sysname\ using four different synthetic generators.  First observe that recent diffusion models like TabSyn and ForestFlow (last two columns) provide higher quality generations as measured by their visual similarity to the original data.  We zoom in further to highlight the difference between baseline generations and \sysname\ generations.  We assign to each generated minority point an error label based on the gold $P(y|x)$ model that is known in this toy setting.   In the second row, we show the generated minority with wrong generations shown in pink from the baseline model.  In the third row, we show the same for \sysname\ variants of the generators.  Observe that there are significantly fewer wrong generations from our \sysname\ generators.  We will later quantify the exact error of generation.

\paragraph{Overall Accuracy Comparison on real datasets}
In this section, we first report the main accuracy results for all synthesizers and how overall accuracy increases due to \sysname in Table \ref{tab:model_performance} and other metrics in Table 3 of the appendix.  From these tables we can make the following important observations.  (1) Our proposed \sysname\ with CTabSyn, a conditional latent diffusion model, as a generator, provides the highest accuracy across all four datasets and all classifiers. (2) Compared to the baseline classifier trained on original data $D$ (first row of numbers), most synthetic data generation methods boost accuracy.  The only exception is Fintech where only our methods provide gains.  (3) If we replace the data generator from CTabSyn with any of the alternative generators, our \sysname\ extension of training the generator and classifier, provides a boost in accuracy. 

A paired T-test was performed on 5 synthesizers and 4 datasets with the balanced test sets, comparing accuracy with and without \sysname. A p-value of 6.84e-05 $<$ 0.05 was obtained, \textbf{showing the statistical significance of \sysname}.

We next investigate the reasons behind the gains that \sysname\ provides over the respective baseline generative model. \sysname\ impacts both the data synthesis step and the classifier training step.  In subsequent experiments, we tease apart the contribution of each step in the overall accuracy gains.  


\xhdr{ORD Gains Are Due to Better Synthetic Data Generation}
We first demonstrate that generators $G$ trained with \sysname's ternary labels provide better quality instances. We wanted to check the agreement in class label assigned by generator and true label of generated instances.  However, the true label is not available for real data.  So, we found two work arounds.  First, for $D$ we used a toy dataset where the true class distribution $P(y|x)$ is known. Table~\ref{tab:toy_mog} shows the results on a mixture of Gaussian dataset as shown in Figure~\ref{fig:toy-vis}.  Observe that in all cases minority accuracy is worse than majority, and recent diffusion based models like TabSyn and ForestFlow provide more accurate generations.  Our extension \sysname\ improves the minority accuracy even further in all cases, and also provides modest improvement for majority instances.  Second, for experiments on real data, we trained an accurate XG Boost model on real balanced data. Note that this data is not available to train the synthesizer, and is only used here to define an oracle.  Table~\ref{tab:randomoracle} shows overall accuracy of generated instances.  Across all generators and datasets, we observe that our \sysname\ variant generates instances with greater accuracy.


\begin{table}[h!]
  \centering
\begin{tabular}{l|p{.15\linewidth}|p{.15\linewidth}|p{.15\linewidth}} \hline
Model                & Avg of Maj, Min (\%) & Minority Acc. (\%) & Majority Acc.(\%) \\\hline\hline
CTabGan+             & 67.98        & 41.67    & 94.30    \\
CTabGan+ \sysname    & 70.52        & 46.08    & 94.95    \\ \hline
DDPM                 & 93.51        & 88.72    & 98.31    \\
DDPM \sysname        & 94.21        & 89.87    & 98.56    \\ \hline
TabSyn               & 91.93        & 85.75    & 98.10    \\
CTabSyn \sysname      & 97.72       & 96.50    & 98.93    \\ \hline
ForestFlow           & 95.68       & 91.90    & 99.47    \\
ForestFlow \sysname  & 95.74        & 92.02    & 99.46    \\\hline
\end{tabular}
  \caption{Accuracy of synthetic data label with true labels coming from the Bayes classifier. Minority data quality improves by the \sysname method in most synthesizers. Since the generated data is mostly majority, Overall accuracy as the weighted avg. of majority and minority accuracies.}
  \label{tab:toy_mog}
\end{table}

\begin{table}[h!]
\centering
\setlength\tabcolsep{2.0pt}
\begin{tabular}{l|cccc}
\hline 
\textbf{Model} & \textbf{Adult} & \textbf{Cardio} & \textbf{Fintech} & \textbf{Heloc} \\
\hline \hline
CTGAN              & 67.94 & 70.89 & 55.22 & 54.84 \\
CTGAN \sysname     & 67.24 & 64.38 & 56.36 & 53.97 \\ \hline
CTAB-GAN+          & 54.54 & 68.08 & 53.69 & 52.84 \\
CTAB-GAN+ \sysname & 77.06 & 70.48 & 50.92 & 49.80 \\\hline
TabDDPM            & 75.38 & 59.35 & 65.83 & 49.80 \\
TabDDPM \sysname   & 77.36 & 73.60 & 61.53 & 73.82 \\\hline
TabSyn             & 67.19 & 69.22 & 60.73 & 63.86 \\
CTabSyn \sysname   & 72.64 & 72.07 & 65.97 & 66.91 \\\hline
ForestFlow             & 69.78 & 73.49 & 77.12 & 74.01 \\
ForestFlow \sysname    & 70.21 & 73.54 & 72.42 & 73.23 \\\hline
\end{tabular}
\caption{Accuracy of testing synthetic data on a classifier trained on real \textbf{balanced} datasets. An increase with \sysname shows how synthesis is improved by \sysname .}
\label{tab:randomoracle}
\end{table}

\subsubsection{\sysname\ Trains a Better Classifier by Downsampling Overlapping Majority}
Note for training the classifier we only sample instances from clear majority and minority.   We show that this is another reason why \sysname\ leads to higher classification accuracy.  We disantangle the effect of the quality of synthetic instances by training classifiers on real dataset.  
We train two classifiers: the first is trained on the entire dataset - $\CM$ and $\Cm$ while second is trained in \sysname\ mode where we remove the instances from $D_0$ that falls in the overlap region.
We show the results in Table~\ref{tab:real_00_1} and observe that overall accuracy increases with removing Overlap majority points in most cases.


\begin{table}[h!]
  \centering
  \begin{tabular}{l|p{0.18\linewidth}p{0.18\linewidth}p{0.18\linewidth}} \hline
    Metric           & Accuracy (\%) & Minority Accuracy (\%) & Majority Accuracy (\%) \\\hline\hline
    Adult            & 76.23         & 57.06                  & 95.40                  \\
    Adult \sysname   & 81.57         & 75.68                  & 87.45                  \\ \hline
    Fintech          & 69.25         & 54.27                  & 85.60                  \\
    Fintech \sysname & 72.97         & 76.77                  & 69.16                  \\ \hline
    Heloc            & 71.04         & 65.95                  & 76.14                  \\
    Heloc \sysname   & 70.55         & 84.56                  & 56.52                  \\ \hline
  \end{tabular}
  \caption{Effect of removing Overlap i.e. $\CmM$ on real data is shown in this table. The improvement shows that removing $\CmM$ helps classifier to learn better decision boundary.}
  \label{tab:real_00_1}
\end{table}

\paragraph{Ablations}  
We perform another interesting ablation to show that \sysname directly affects the synthetic data generator and not just the classifier. Here we compare \sysname\ with an ablation where we directly use $D$ to train the generators, sample  $\CM$ and $\Cm$ from them, and then remove overlap instances from $\CM$.  Table~\ref{tab:assistSyn} presents a comparison of accuracy of the XGBoost classifier trained using the synthetic data in the two settings.  We observe that for  two of the datasets --- Adult and Heloc, the accuracy drops by a huge amount when the filtering of overlapping instances is done after training the generator on binary labels.  This is another evidence to support our claim that the finer-grained ternary labels leads to a better quality generator.   In Table~\ref{tab:model_performance} we did not include the real minority instances to train the classifier to enable fair comparison across all methods. We have found that, out of the various possible augmentation combinations, augmenting the synthetic data with only the minority class $\Cm$ original data points provides the greatest gains. The results are tabulated in Table \ref{tab:augmentation}.   We present more ablations in the supplementary.

\begin{table}[h]
\centering
\begin{tabular}{l|p{0.3\linewidth}|p{0.3\linewidth}}
  \hline
  Dataset & \sysname Before Synthesis (\%) & \sysname After Synthesis (\%) \\ \hline\hline
  Adult            & 79.19                          & 65.32                         \\
  Heloc            & 67.00                          & 44.72                         \\
  Fintech          & 60.19                          & 60.90                         \\
  Cardio            &  72.04                              & 72.31                        \\\hline
\end{tabular}
\caption{Comparison of XGBoost Accuracy of \sysname done Before and After Synthesis shows that \sysname aids Synthesis. It shows that not just removing overlap for the classifier but also providing that information to the synthesizer is necessary.}
\label{tab:assistSyn}
\end{table}

\begin{table}[h!]
\setlength\tabcolsep{2.0pt}
\centering
\begin{tabular}{p{0.39\linewidth}|p{.12\linewidth} p{.12\linewidth}p{.12\linewidth} p{.12\linewidth}}
\hline
Training Data & Adult (\%) & Fintech (\%) & Heloc (\%) & Cardio (\%) \\ \hline\hline
Real Unbalanced & 46.52 & 25.87 & 31.04  &  32.46\\ 
Synthetic \sysname & 78.30 & 60.52 & 65.74 & 63.81 \\
Synthetic \sysname + $D_1$ & 79.15 & 63.92 & 67.40 & 65.62\\
\hline
\end{tabular}
\caption{Effect of augmenting synthetic data with real minority data $D1$.  There is an improvement in classification accuracy. The Classifier used is an  Average of 4 classifiers LR, MLP, DT and Adaboost.}
\label{tab:augmentation}
\end{table}

\section{Conclusion, Limitations, Future Work}
In this paper, we proposed \sysname, a new technique for training classifiers with highly imbalanced class distribution for augmenting with synthetic data from generative models.  A key idea of \sysname\ is identifying overlapping majority instances and converting the original binary labels into a ternary labeled dataset. We show that existing deep generative models are also adversely affected by imbalance in training data, and show that class conditional generators trained with our ternary labels provide higher quality data.  Finally, \sysname\ under-samples overlapping majority instances and trains the classifier using balanced synthetic minority and clear majority instances along with real minorities.  We present a detailed comparison on four real datasets and obtain improvement in classification accuracy in most settings.   We explain the reasons behind the gains via three insightful experiments that show that \sysname\ improves both the data synthesis step and the classifier training step.

\paragraph{Limitations and Future Works}
\begin{enumerate*}[(1)]
    \item The method proposed is for datasets having binary class target columns only. However, with some tweaking, there is a possibility of extending this method to datasets with multi-class categorical target columns.
    \item While the method has been shown to work on datasets with highly imbalanced categorical columns, it will be interesting to extend to datasets with continuous-valued target columns with a very skewed distribution.
    \item The focus of this work is not privacy preservation but an interesting extension is to train the data generator without compromising privacy of the real data. 
\end{enumerate*}

\section*{Acknowledgments}
 Work done in this project was funded by the State Bank of India (SBI) Data Analytics Hub at IIT Bombay.  We thank SBI data scientists for introducing us to the problem.  We acknowledge discussions with Dr Rajbabu in initial phases of the project. 

\bigskip
\noindent 
\bibliography{aaai25}
\appendix

\include{appendix}
\end{document}

%% file: appendix.tex
\title{Supplementary material}
\section{System Specifications and Complexity Analysis}
\subsection{Hardware and Software used}
\textbf{GPU/CPU Models} \\
\textbf{CPU:} Intel(R) Xeon(R) Platinum 8160 CPU @ 2.10GHz \\
\textbf{GPU:} NVIDIA RTX A6000\\
\textbf{Memory} \\
\textbf{CPU:} 1.5 TiB \\
\textbf{GPU:} 49140 MiB \\
\textbf{Operating System:} Linux\\
\textbf{Software Libraries and Frameworks}
\begin{tabbing}
\hspace{4cm} \= \kill
pandas \> 2.2.2 \\
numpy \> 2.0.0 \\
matplotlib \> 3.9.0 \\
scikit-learn \> 1.5.0 \\
scipy \> 1.13.1 \\
sdmetrics \> 0.14.1 \\
\end{tabbing}
\subsection{Complexity Analysis}
 The primary computation in \sysname is training an ensemble of classifiers and obtaining confidence scores. The complexity of the synthetic data generation depends on the deep generative model used (not part of \sysname). We have tried real-world datasets with a large number of features and samples. Our method scales well even upto 200 features and dataset sizes upto 250,000 taking 900s and 350s respectively. Details tabulated in the Table \ref{tab:increasing_features} and Table \ref{tab:increasing_dataset_size}.
 
 \begin{table}[h!]
  \centering
\begin{tabular}{r|r|r} \hline
\textbf{num samples} & \textbf{num features} & \textbf{Time (s)} \\ \hline
50000               & 20                   & 90.473581         \\ 
50000               & 50                   & 164.960408        \\ 
50000               & 100                  & 246.302797        \\ 
50000               & 200                  & 351.647613        \\ \hline
\end{tabular}
  \caption{Execution time for \sysname preprocessing increasing number of features while keeping the number of samples fixed at 50,000.}
  \label{tab:increasing_features}
\end{table}

\begin{table}[h!]
  \centering
\begin{tabular}{r|r|r} \hline
\textbf{num samples} & \textbf{num features} & \textbf{Time (s)} \\ \hline
10000               & 30                   & 18.172544         \\ 
50000               & 30                   & 115.197481        \\ 
100000              & 30                   & 257.074255        \\ 
200000              & 30                   & 568.373154        \\ 
300000              & 30                   & 911.999833        \\ \hline
\end{tabular}
  \caption{Execution time for \sysname preprocessing increasing dataset size while keeping the number of features fixed at 30.}
  \label{tab:increasing_dataset_size}
\end{table}

\begin{table*}[h!]
  \centering
\begin{tabular}{l|l|rr rr rr} \hline
Dataset & Synthesizer & Avg of 4 & XGBoost & Min Acc (\%) & Maj Acc (\%) & F1 & AUCROC \\\hline\hline
Abalone & CTAB-GAN+       & \textbf{23.21} & \textbf{63.16} & 62.86 & \textbf{63.19} & 22.22 & 67.96 \\
        & CTAB-GAN+ ORD   & 23.11 & 56.10 & \textbf{80.00} & 53.92 & \textbf{23.38} & \textbf{74.03} \\
        & ForestFlow         & \textbf{39.34} & \textbf{83.97} & 61.43 & \textbf{86.03} & 39.09 & 84.16 \\
        & ForestFlow ORD     & 38.65 & 82.18 & 67.14 & 83.55 & \textbf{38.68} & \textbf{84.57} \\\hline
Bank    & CTAB-GAN+       & 45.71 & \textbf{82.61} & 62.14 & \textbf{85.41} & 46.30 & 83.10 \\
        & CTAB-GAN+ ORD   & \textbf{84.45} & 79.61 & \textbf{75.16} & 80.22 & \textbf{47.07} & \textbf{85.36} \\
        & ForestFlow         & 50.02 & 88.29 & \textbf{74.61} & 90.17 & 60.49 & \textbf{92.06} \\
        & ForestFlow ORD     & \textbf{51.30} & \textbf{88.78} & 70.85 & \textbf{91.23} & \textbf{60.37} & 91.66 \\\hline
Car     & CTAB-GAN+       & 7.02  & 52.60 & 54.55 & 52.54 & 6.82  & 55.01 \\
        & CTAB-GAN+ ORD   & \textbf{13.64} &\textbf{ 61.27} & \textbf{100.00} & \textbf{60.00} & \textbf{14.10} & \textbf{84.15} \\
        & ForestFlow         & 37.42 & 95.38 & \textbf{100.00} & 95.22 & 57.89 & 98.56 \\
        & ForestFlow ORD     & \textbf{38.59} & \textbf{96.53} & \textbf{100.00} & \textbf{96.42} & \textbf{64.70} & \textbf{98.67} \\\hline
Yeast   & CTAB-GAN+       & 33.76 & 52.91 & 38.85 & \textbf{72.22} & 23.53 & 57.01 \\
        & CTAB-GAN+ ORD   & \textbf{39.14} & \textbf{55.07} & \textbf{86.11} & 50.77 & \textbf{31.79} & \textbf{74.76} \\
        & ForestFlow         & 75.99 & \textbf{93.58} & 80.56 & \textbf{95.38} & 75.30 & \textbf{97.00} \\
        & ForestFlow ORD     & \textbf{76.00} & 92.23 & \textbf{86.11} & 93.08 & \textbf{76.94} & 96.36 \\\hline
\end{tabular}

  \caption{Metrics for UCI datasets Abalone, Bank, Car and Yeast include F1 score and AUC since the test sets are imbalanced. The minority increase is considerable in all datasets while F1 improves in with \sysname in all cases while AUC shows an improvement in 6 out of 8 cases.}
  \label{tab:uci_results}
\end{table*}

\section{Measures of variation for the main table}
For the table computing machine learning efficacy for 4 datasets on 5 synthesizers with and without \sysname, we compute the standard deviation over 3 runs sampling with multiple seeds during each run. The results are tabulated in Table \ref{tab:sd-mle}.
\begin{table*}[h!]
    \centering
    \begin{tabular}{l|rr|rr|rr|rr}
        \hline
        & \multicolumn{2}{c|}{AI} & \multicolumn{2}{c|}{Cardio} & \multicolumn{2}{c|}{Fintech} & \multicolumn{2}{c}{Heloc} \\ \cline{2-9}
        Model & Avg of 4 & XGBoost & Avg of 4 & XGBoost & Avg of 4 & XGBoost & Avg of 4 & XGBoost \\ \hline\hline
        CTGAN & 0.01 & 0.73 & 0.01 & 1.35 & 0.01 & 1.11 & 0.01 & 0.74 \\
        CTGAN \sysname & 0.01 & 1.08 & 0.00 & 1.40 & 0.01 & 0.89 & 0.01 & 0.56 \\ \hline
        CTAB-GAN+ & 0.65 & 0.44 & 0.19 & 2.11 & 1.07 & 0.25 & 0.66 & 1.45 \\
        CTAB-GAN+ \sysname & 0.80 & 0.95 & 0.80 & 2.02 & 0.90 & 2.14 & 1.17 & 0.92 \\ \hline
        Forest & 0.53 & 1.28 & 0.25 & 0.73 & 1.50 & 0.13 & 1.34 & 0.87 \\
        Forest \sysname & 1.13 & 1.35 & 0.58 & 0.88 & 0.13 & 0.84 & 1.89 & 1.04 \\ \hline
        TabDDPM & 0.09 & 0.95 & 0.02 & 1.00 & 0.10 & 0.90 & 0.01 & 1.08 \\
        TabDDPM \sysname & 0.02 & 0.18 & 0.02 & 0.33 & 0.03 & 0.27 & 0.07 & 0.64 \\ \hline
        TabSyn & 0.36 & 2.15 & 0.84 & 1.19 & 0.48 & 0.61 & 0.61 & 0.59 \\
        CTabSyn & 0.58 & 0.27 & 0.68 & 0.36 & 0.39 & 0.34 & 0.40 & 0.60 \\
        CTabSyn \sysname & 0.25 & 0.12 & 0.65 & 0.36 & 0.26 & 0.36 & 0.58 & 0.33 \\ \hline
    \end{tabular}
     \caption{For Machine learning efficacy (accuracy of classifier trained with synthetic data) reported in main paper, the following is the Standard deviation over 3 runs.}
  \label{tab:sd-mle}
\end{table*}

\section{Ablation study and further experiments}
\subsection{Not a trivial boundary shifter}
Since we showed that the method of conditioning creates a decluttered boundary for the minority class points, one might think that this could be synonymous with shifting the classification boundary towards the majority class examples, thus increasing minority class accuracy while reducing majority class accuracy. We strike out this assumption with the following experiment.

We split the original dataset D into train and test sets, $D_{train}$ and $D_{test}$, respectively. We train an XGBoost classifier \textit{Clf} on $D_{train}$ and obtain the class probability values on the test
set $D_{test}$. Now we obtain the scores on the test set using different values of thresholds (essentially lowering it so that the classification boundary shifts more towards the majority class).  We find the  threshold for which we get the best scores and compare these scores with the scores obtained by our method of \sysname and observe that in most cases, the method of \textbf{\sysname outperforms the shifting threshold method}. Thus, this shows that \sysname is non-trivial and is not analogous to simply shifting the classifier boundary. The results are tabulated in Table \ref{tab:boundaryShifter}.\\
\begin{table}[h]
	\centering
	\begin{tabular}{l|p{0.2\linewidth}|p{0.22\linewidth}|p{0.18\linewidth}}
		\hline
		Dataset               & Score at 0.5 (\%) & Best Threshold & Best Score (\%) \\ \hline\hline
		Adult                 & 80.12             & 0.48           & 80.25           \\
		Adult  \sysname       & 80.60             & 0.45           & 81.02           \\
		Fintech               & 60.75             & 0.45           & 61.70           \\
		Fintech  \sysname     & 62.20             & 0.45           & 62.64           \\
		Heloc                 & 44.82             & 0.45           & 66.45           \\
		Heloc   \sysname      & 66.45             & 0.50              & 66.45               \\ 
            Cardio                 & 72.52             & 0.49           & 72.58           \\
		  Cardio  \sysname      & 72.75             & 0.50              & 72.75              \\\hline
	\end{tabular}
	\caption{Results show that best threshold for \sysname is always better than best for that of no \sysname. \sysname is hence not a trivial boundary shifter.}
	\label{tab:boundaryShifter}
\end{table}

\subsection{Augmentation for further improvement}
When privacy is not a concern, the original data augmented with synthetic data results in the training of a better classifier. We have found that, out of the various possible augmentation combinations, augmenting the synthetic data with only the minority class $\Cm$ original data points improves machine learning efficacy (MLE). The results are tabulated in Table \ref{tab:augment-adult} and Table \ref{tab:augment-cardio}.

\section{Experimenting with different thresholds}
\subsection{Number of overlapped instances for different thresholds}
\begin{figure}
    \centering
    \includegraphics[width=1\linewidth]{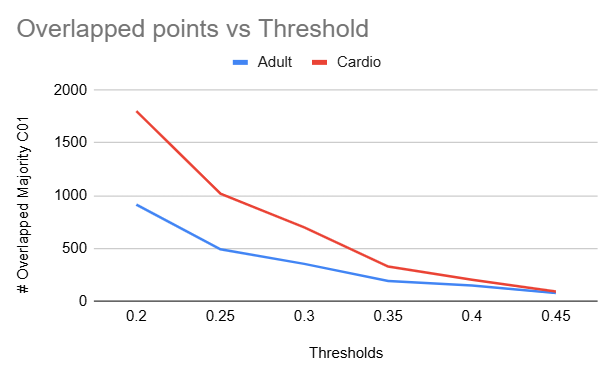}
    \caption{Varying thresholds changes the number of $\CmM$ points in the dataset after applying overlap detection. The exact numbers are reported in Table \ref{tab:thres-table1}}
    \label{fig:thres-graph1}
\end{figure}

\begin{table}[h!]
\centering
\begin{tabular}{r|r|r}
\hline
Threshold& Adult & Cardio \\ \hline\hline
0.20       & 915            & 1801            \\ 
0.25      & 493            & 1019            \\ 
0.30       & 355            & 700             \\ 
0.35      & 192            & 330             \\ 
0.40       & 151            & 205             \\ 
0.45      & 79             & 93              \\ \hline
\end{tabular}

\caption{The effect of varying thresholds on the number of majority points in the overlap region. Even though the original dataset has a similar sizes, the difference in overlapped points for the same threshold indicates the intrinsic difference in the overlap region of the original dataset as shown in Figure \ref{fig:thres-graph1}}
\label{tab:thres-table1}
\end{table}
It is obvious that when the threshold $\tau$ is high of the order of 0.4 or 0.45, we get very little overlapped class $\CmM$ points. But on the other hand, if it is too low like 0.15 or 0.2, only the instances with Majority confidence above 0.8 or 0.85 get into Clear Majority $\CMM$. This relationship is demonstrated for Adult and Cardio datasets by the Table \ref{tab:thres-table1} and shown in the Figure \ref{fig:thres-graph1}

\subsection{Machine Learning Efficacy varying with thresholds}
We converted the binary target to ternary label dataset with different thresholds ranging from 0.2 to 0.45 and then repeated the MLE experiment. It is found that different thresholds are best suited for different datasets. Use of Validation data for best threshold decision was discussed in main paper. A threshold of 0.3 or 0.35 works best for Adult and Cardio with extreme imbalance considered. Further study on the Machine Learning Efficacy is presented in Table \ref{tab:thres-table2} for data Adult and Table \ref{tab:thres-table3} for data Cardio.

\begin{figure}
    \centering
    \includegraphics[width=1\linewidth]{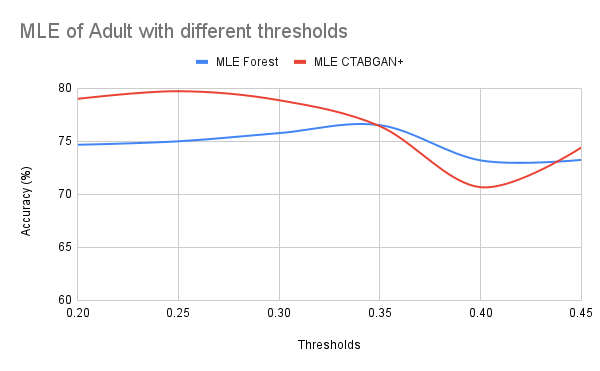}
    \caption{MLE of dataset Adult with varying thresholds during \sysname}
    \label{fig:thres-graph2}
\end{figure}

\begin{figure}
    \centering
    \includegraphics[width=1\linewidth]{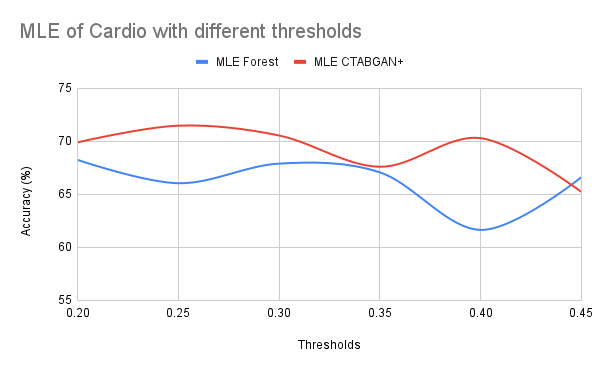}
    \caption{MLE of dataset cardio with varying thresholds during \sysname}
    \label{fig:thres-graph3}
\end{figure}

Observations are that, Machine learning efficacy improves until an optimal threshold and then decreases again. There are instances where more than one threshold is suitable for a given dataset.

\begin{table*}[ht]
    \centering
    \begin{tabular}{l|llll|llll}
        \hline
        & \multicolumn{4}{|c|}{\textbf{Adult, CTABGAN+}} &  \multicolumn{4}{c}{\textbf{Adult, ForestFlow}} \\
         Threshold &  \textbf{Avg of 4} & \textbf{XGBoost} & \textbf{Min acc} & \textbf{Maj acc}  & \textbf{Avg of 4} & \textbf{XGBoost} & \textbf{Min acc} & \textbf{Maj acc} \\
        \hline
        0.20 & 74.63 & 79.00 & 73.00 & 85.00 & 73.24 & 74.68 & 64.30 & 85.05 \\
        0.25 & 79.06 & 79.73 & 76.80 & 82.65 & 67.20 & 75.00 & 69.45 & 80.55 \\
        \textbf{0.30} & 78.69 & 78.88 & 73.65 & 84.10 & 67.87 & 75.78 & 67.15 & 84.40 \\
        0.35 & 75.06 & 76.43 & 66.20 & 86.65 & 0.70 & 76.53 & 72.10 & 80.95 \\
        0.40 & 72.30 & 70.68 & 76.05 & 65.30 & 68.41 & 73.20 & 61.20 & 85.20 \\
        0.45 & 70.63 & 74.43 & 69.30 & 79.55 & 69.80 & 73.25 & 66.65 & 79.85 \\
        \hline
    \end{tabular}
    
\caption{Varying Thresholds while detecting overlap and measuring MLE for Adult data. Machine Learning efficacy experiment with Average of 4 classifiers LR, MLP, Adaboost and DT from SD Metrics along with XGBoost classifier. XGboost classifier's minority and majority accuracy is also noted to see if minority increases well. Found 0.3 to be the best threshold for Adult dataset.}
\label{tab:thres-table2}
\end{table*}

\begin{table*}[ht]
    \centering
    \begin{tabular}{l|llll|llll}
        \hline
        & \multicolumn{4}{|c|}{\textbf{Cardio, CTABGAN+}} &  \multicolumn{4}{c}{\textbf{Cardio, ForestFlow}} \\
         Threshold &  \textbf{Avg of 4} & \textbf{XGBoost} & \textbf{Min acc} & \textbf{Maj acc}  & \textbf{Avg of 4} & \textbf{XGBoost} & \textbf{Min acc} & \textbf{Maj acc} \\
        \hline
         0.20 & 72.16 & 69.90 & 80.28 & 59.49 & 67.39 & 68.25 & 61.16 & 75.36 \\
        0.25 & 71.81 & 71.48 & 76.29 & 66.65 & 67.04 & 66.05 & 60.61 & 71.51 \\
        0.30 & 71.01 & 70.55 & 74.54 & 66.55 & 67.21 & 67.90 & 63.21 & 72.61 \\
        \textbf{0.35} & 71.17 & 67.60 & 80.58 & 54.58 & 66.28 & 67.07 & 58.36 & 75.81 \\
        0.40 & 71.37 & 70.30 & 79.58 & 60.99 & 65.62 & 61.65 & 71.59 & 51.68 \\
        0.45 & 70.28 & 65.23 & 87.47 & 42.91 & 65.80 & 66.63 & 58.76 & 74.51 \\
        \hline
        \hline
    \end{tabular}
    
\caption{Varying Thresholds while detecting overlap and measuring MLE for Cardio data. Machine Learning efficacy experiment with Average of 4 classifiers LR, MLP, Adaboost and DT from SD Metrics and XGBoost classifier. XGboost classifier's minority and majority accuracy is also noted to see if minority increases well. Found lower thresholds than 0.35 to be the optimal ones for Cardio dataset.}
\label{tab:thres-table3}
\end{table*}
\begin{table*}
\begin{tabular}{c|c|c|c|c|c|c|c|c}
& \multicolumn{4}{|c|}{\textbf{Adult, CTABGAN+}} &  \multicolumn{4}{c}{\textbf{Adult, ForestFlow}} \\
\textbf{Data} & \textbf{Avg of 4} & \textbf{XGboost} & \textbf{Min acc} & \textbf{Maj acc} &  \textbf{Avg of 4} & \textbf{XGboost} & \textbf{Min acc} & \textbf{Maj acc} \\
\hline \hline
$S_{00} \cup S_1$                               & 78.96 & 79.73 & 76.80 & 82.65 &  67.46 & 75.00 & 69.45 & 80.55 \\
$S_{00} \cup S_1 \cup \text{subsample}(S_{01})$ &74.67 & 76.33 & 66.40 & 86.25 &  63.67 & 69.27 & 53.35 & 85.20 \\
$S_{00} \cup S_1 \cup D_1$                      & 79.57 & 81.18 & 80.35 & 82.00 &  78.08 & 70.45 & 98.50 & 42.40 \\
$S_{00} \cup S_1 \cup D_1 \cup \text{subsample}(D_0)$ & 79.83 & 81.83 & 78.15 & 85.50 &  77.38 & 84.23 & 74.80 & 93.65 \\
\hline
\hline
\end{tabular}
\caption{ The results of 2 synthesizers - CTabGAN+ and ForestFlow are shown on the Adult Datset. Here, we see that the best minority accuracy without a loss in majority accuracy is obtained in the case where the synthetic data is augmented with real minority class data. While augmenting synthetic data with real minority class data and some portion of real majority class data gives the best overall accuracy (though not necessarily the best minority class accuracy)}
\label{tab:augment-adult}
\end{table*}

\begin{table*}
\begin{tabular}{c|c|c|c|c|c|c|c|c}
& \multicolumn{4}{|c|}{\textbf{Cardio, CTABGAN+}} &  \multicolumn{4}{c}{\textbf{Cardio, ForestFlow}} \\
\textbf{Data} & \textbf{Avg of 4} & \textbf{XGboost} & \textbf{Min acc} & \textbf{Maj acc} &  \textbf{Avg of 4} & \textbf{XGboost} & \textbf{Min acc} & \textbf{Maj acc} \\
\hline \hline
$S_{00} \cup S_1$                               & 70.27	&72.32&	67.15&	77.50&68.62&	67.15&	69.75&	64.55\\
$S_{00} \cup S_1 \cup \text{subsample}(S_{01})$ & 67.39	&70.35&	58.40&	82.30&63.30&	60.22&	55.00&	65.45\\
$S_{00} \cup S_1 \cup D_1$                      & 79.58	&81.18&	80.35&	82.00&69.60&	67.04&	78.82&	55.25\\
$S_{00} \cup S_1 \cup D_1 \cup \text{subsample}(D_0)$ & 79.83	&81.83&	78.15&	85.50&72.00&	80.90&	66.60&	95.20\\
\hline
\end{tabular}
\caption{The results of 2 synthesizers - CTabGAN+ and ForestFlow are shown on the Cardio Datset. The conclusions of this table are the same as those in Table \ref{tab:augment-adult}}
\label{tab:augment-cardio}
\end{table*}

\section{Experiment with 2D Toy}
\textit{make\_blobs} from the sklearn.datasets library was used for creating the toy dataset. $num_{maj}$ points were equally distributed among $n_{maj}$ blobs while $num_{min}$ points were equally distributed among $n_{min}$ blobs. $num_{maj}$, $num_{min}$, $n_{maj}$, $n_{min}$, the centres and standard deviations of all the blobs have been provided in the Table \ref{tab:toy_dataset_details}.

\begin{table}[h!]
  \centering
  \begin{tabular}{l|l|l|p{.2\linewidth}|l}
    \hline
    Data & \#blobs & num & Centers & Std. Dev. \\ \hline\hline
    Majority 1   & 4      & 8K            & [-30,-30], [20,40], [0,0], [40,-50] & [3, 4, 10, 6] \\
    Minority 1    & 3      & 0.5K           & [0,-30], [25,20], [20,-50]  & [4, 6, 4] \\ \hline
    Majority 2   & 3     & 7.5K            & [10,10], [-30,-20], [40,-50] & [10, 8, 4] \\
    Minority 2   & 2     & 0.2K            & [-30,10], [20,-20] & [10,4 ] \\ \hline
  \end{tabular}
    \caption{2 different toy datasets are created with differing number of majority and minority blobs. First toy has been tabulated in the main paper while second toy results are present in Table \ref{tab:toy_1}.}
  \label{tab:toy_dataset_details}
\end{table}

\begin{figure}
    \centering
    \includegraphics[width=1\linewidth]{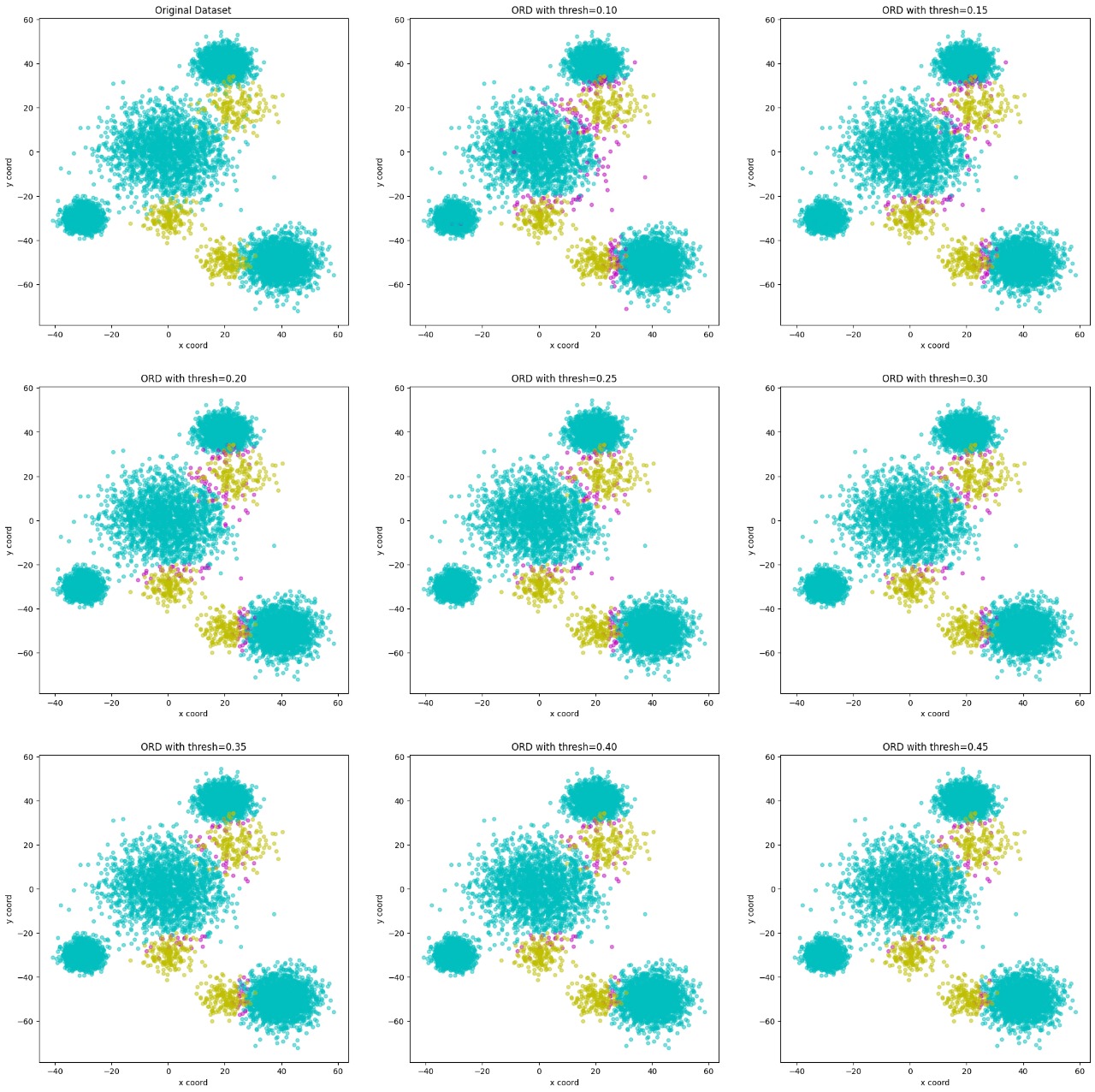}
     \caption{Toy 2 visusalisation showing \sysname on different thresholds and the overlapped majority in a different color. }
    \label{fig:toy-thres}
\end{figure}

\begin{figure*}
    \centering
    \includegraphics[width=1\linewidth,,height=3.8in]{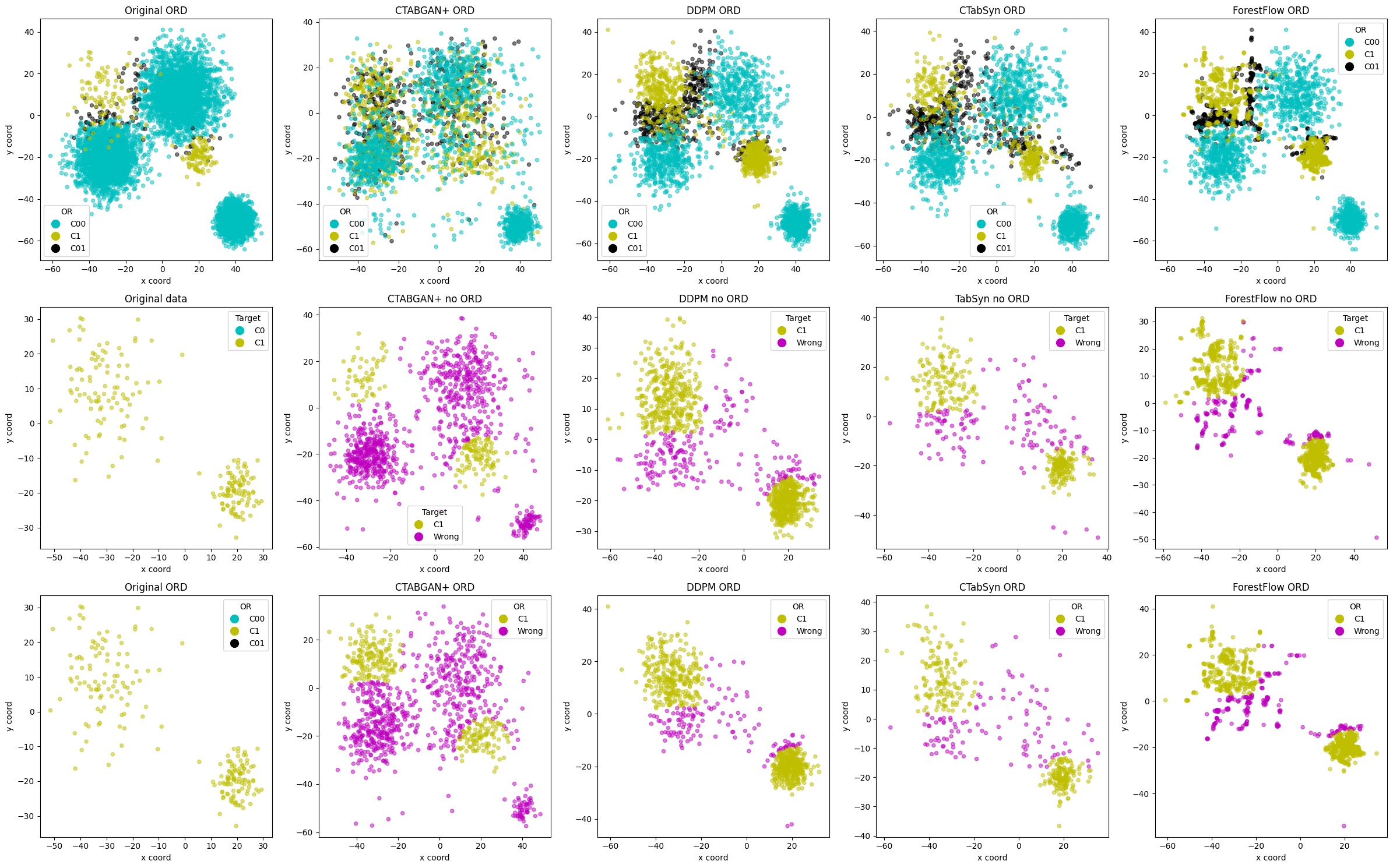}
     \caption{Toy 2 visusalisation}
    \label{fig:toy-2-vis}
\end{figure*}

A visualisation of toy 1 is present in main paper. The visualisation of toy 2 is presented in Figure \ref{fig:toy-2-vis}. A visualisation of the toy for different thresholds showing different boundary points is shown in Figure \ref{fig:toy-thres} 

Since the centers and standard deviation of the blobs are known, we approximate the accuracy of the synthetic data using simple Naive Bayes using a Mixture of Gaussians (MoG) of the blobs of the same class.
\begin{equation}
\begin{aligned}
& P(y=1|X) = P(X|y=1)P(y=1) / P(X) \\
& P(y=0|X) = P(X|y=0)P(y=0) / P(X) \\
& f(X) =
\begin{cases}
1 & \text{if } P(y=1|X) > P(y=0|X) \\
0 & \text{otherwise}
\end{cases}
\end{aligned}
\end{equation}


\begin{table}[h!]
  \centering
\begin{tabular}{l|p{.15\linewidth}|p{.15\linewidth}|p{.15\linewidth}} \hline
Model                & Avg of Maj, Min (\%) & Minority Acc. (\%) & Majority Acc.(\%) \\\hline\hline
CTabGan+             & 54.48&	15.25&	93.70    \\
CTabGan+ \sysname    & 61.55&	27.42&	95.67    \\ \hline
DDPM                 & 89.66	&79.68&	99.64   \\
DDPM \sysname        & 91.42&	83.16&	99.68    \\ \hline
TabSyn               & 85.63&	71.59&	99.67    \\
CTabSyn \sysname      & 86.21&	73.02&	99.4    \\ \hline
ForestFlow           & 87.02&	74.4&	99.64    \\
ForestFlow \sysname  & 89.33&	78.87&	99.78   \\\hline
\end{tabular}
  \caption{Accuracy of synthetic data (Blobs2) label with true labels from the Bayes classifier. Minority data quality improves by the \sysname method in most synthesizers.}
  \label{tab:toy_1}
\end{table}

\section{Experiment with 1D Toy}
\subsection{Experimentation with 1D gaussian}
\begin{figure}
    \centering
    \includegraphics[width=0.5\linewidth]{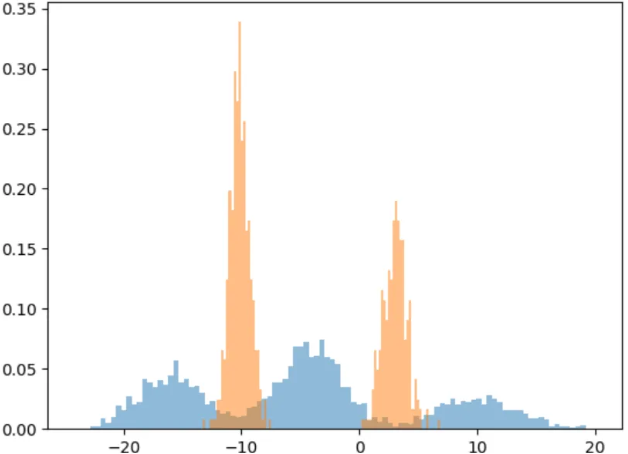}
    \caption{\small The distribution of the 1D mixture of gaussians}
    \label{fig:gauss1}
\end{figure}

\begin{figure}
    \centering
    \includegraphics[width=0.5\linewidth]{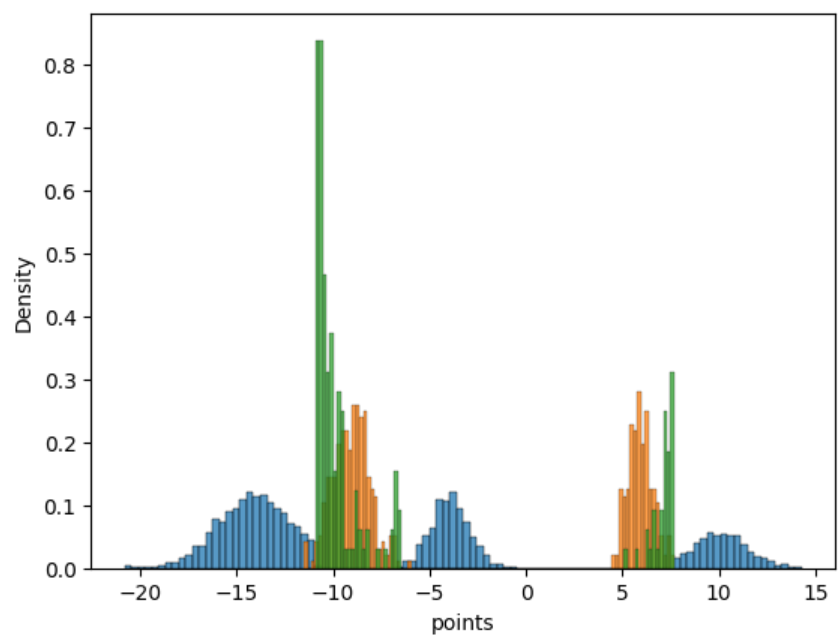}
    \caption{\small The distribution of the 1D mixture of Gaussians with the overlap majority class detected}
    \label{fig:gauss2}
\end{figure}

\begin{figure*}
    \centering
    \includegraphics[width=0.5\linewidth]{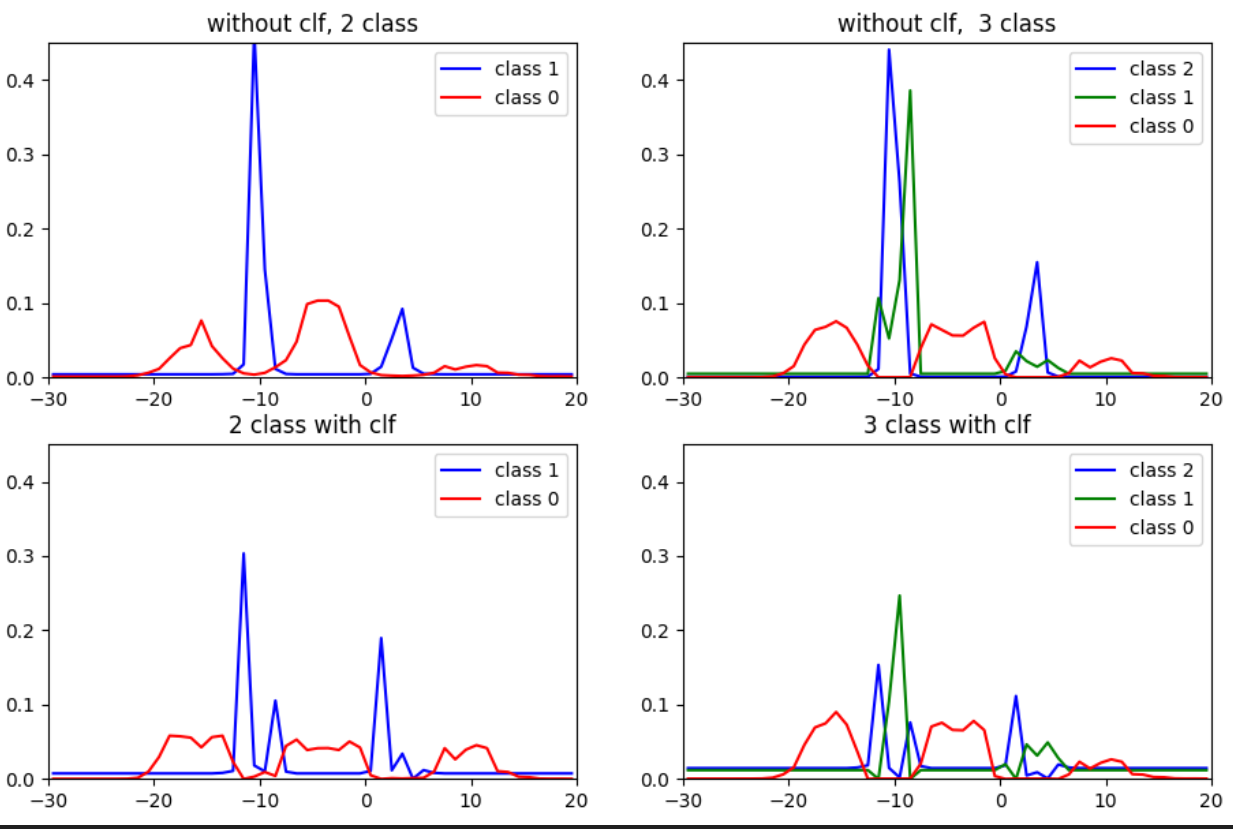}
    \caption{\small The outputs of using 2 classes and 3 classes and with and without a classifier loss}
    \label{fig:gauss3}
\end{figure*}

The following experiment was conducted to demonstrate why the ORD method works i.e. why adding information about a third (overlapped majority) class benefits the minority class and enables us to learn a better distribution for it.
The experiment was as follows:
\begin{itemize}
\item Created a 1D dataset consisting of 2 classes (majority and minority) sampled from mixture of gaussians. The densities of the minority and majority classes created have been given in Figure \ref{fig:gauss1}.
\item For the given distribution, the ORD points i.e. the overlapping majority points were detected using Naive Bayes probability of the majority class belonging to the majority and the minority classes. The resulting density has been displayed in Figure \ref{fig:gauss2}.
\begin{equation}
\begin{aligned}
& P(y=1|X) = P(X|y=1)P(y=1) / P(X) \\
& P(y=0|X) = P(X|y=0)P(y=0) / P(X) \\
& f(X) =
\begin{cases}
1 & \text{if } P(y=1|X) > P(y=0|X) \\
0 & \text{otherwise}
\end{cases}
\end{aligned}
\end{equation}
\item For learning the distributions of the two class as well as the three class case, the X axis of the distribution is divided into 100 bins (hyper parameter), and the classes of the centers of each bin are learnt i.e. the probabilities of the center of each bin belonging to the majority/ minority class is learnt.
\item To show the effect of including a classifier loss in learning the distribution, we first train a logistic regression model on all the input points and their true labels separately for the two-class and three-class cases.
\item Then, we include a term of the loss of the logistic regression classifier when learning the probabilities of the centers of each bin.
\end{itemize}
The learned distributions of each of the cases above have been displayed in figure \ref{fig:gauss3}.
The main observations are as follows:
\begin{itemize}
\item Even in the absence of a classifier, when three classes are used instead of two, it can be observed that the majority distribution is pushed away from the minority distribution
\item Upon using a classifier, irrespective of the 2 classes or 3 classes case, the probability of the minority class in regions away from the bulk minority increases.
\item Upon using three classes and a classifier loss, we see that the majority class is pushed away more than in the case of not using classifier loss. Also, the increase in the minority class probability in regions away from the minority bulk is more in the three classes case.
\end{itemize}
Thus, the above experiment demonstrates the effectiveness of the ORD method in models that contain a classifier loss.

\subsubsection{Preliminary Results}
\begin{itemize}
    \item We observe that when a constraint is given during sampling, irrespective of the number of columns constrained on or whether they are categorical or numerical, the generated samples strictly follow the constraints. Thus, constraint success rate is 100\%
    \item To check the quality of our constrained sampled data, we use SDMetrics to check the \textit{column shapes} and \textit{column pair relations} between the real dataset filtered on those constraints and sampled data 1) with constraints and 2) without constraints filtered on the constraints. Refer to \ref{constraints} for the numbers. It can be observed that in most cases, the constrained data has equal to or better values than the unconstrained data with having slightly worse values in very few cases.
\end{itemize}

\section{Real Datasets}
Eight popular datasets were used for experimental purposes. First four datasets namely - Adult, Cardio (Cardiovascular Disease dataset), HELOC (Home Equity Line of Credit), and Fintech Users. The Adult dataset was taken from UCI Machine Learning Repository\footnote{https://archive.ics.uci.edu/datasets} while the other three datasets were taken from Kaggle\footnote{https://www.kaggle.com/datasets}. The next four are UCI imbalanced datasets - Abalone, Yeast, Car Evaluation and Bank Marketing. The statistics of these datasets have been presented in Table \ref{tab:dataset_characteristics1}.

    
\begin{table}[h!]
  \centering
  \begin{tabular}{l|rrrrr}
    \hline
    Dataset & \# Rows & \# Cat & \# Num & Min/Maj\% \\ \hline\hline
    Adult   & 34514   & 8      & 5            & 1.5  \\
    Heloc   & 5559    & 7      & 16           & 1.9  \\
    Fintech & 11560   & 19     & 9            & 1.8  \\
    Cardio  & 35700   & 6      & 5            & 2  \\ 
   Abalone &  4177  &   1   &    6        &  8.6 \\
    Bank &  45211  &  8    &      6      &  13.1 \\
    Car &  1728  &    5  &       -     &  4.4 \\
    Yeast &  1484  &  1    &      6      & 12\\ 
    
  \end{tabular}
  \caption{Dataset Characteristics. The \#Cat column denotes the number of categorical columns, and \#Real denotes the number of real columns.  In each case, we used 2000 unseen examples each from majority and minority for testing. The percentage of minority points considered for each dataset is between 1.4\%-2\%}
  \label{tab:dataset_characteristics}
\end{table}
The detailed descriptions for each data set are as follows:\\

\textbf{Adult}\footnote{https://archive.ics.uci.edu/dataset/2/adult}: This dataset contains demographic and employment-related features of people \cite{zhang2024mixedtype}. The objective is predicting whether income exceeds \$50K/yr based on census data. This dataset is also known as "Census Income" dataset.

\textbf{Cardio}\footnote{https://www.kaggle.com/datasets/sulianova/cardiovascular-disease-dataset } \textbf{(Cardiovascular Disease dataset)}: This dataset contains objective information i.e. factual information, subjective information i.e. information given by patients  and results of medical examinations. 

\textbf{Fintech}\footnote{https://www.kaggle.com/datasets/niketdheeryan/fintech-users-data}: This dataset contains the financial and demographic data of fintech company users, to improve the sales of the company. The objective of this dataset is classifying whether a customer will purchase something next time or not, also known as customer churn prediction. The objective of this dataset is to predict the presence or absence of disease.

\textbf{HELOC}\footnote{https://www.kaggle.com/datasets/averkiyoliabev/home-equity-line-of-creditheloc} \textbf{(Home Equity Line of Credit)}:  Each entry in the dataset is a line of credit, typically offered by a bank as a percentage of home equity (the difference between the current market value of a home and its purchase price). The objective of this dataset is to predict whether a given line of credit is at risk or not.

\textbf{Yeast} \footnote{https://archive.ics.uci.edu/dataset/110/yeast}: The dataset target variable predicts the cellular localisation sites of proteins. It is a classification dataset in the Biology domain, a popular imbalanced dataset from UCI.

\textbf{Abalone} \footnote{https://archive.ics.uci.edu/dataset/1/abalone}  In the dataset, the age of abalone is determined by cutting the shell through the cone, staining it, and counting the number of rings through a microscope -- a boring and time-consuming task.  Other measurements, which are easier to obtain, are used to predict the age. 

\textbf{Bank} \footnote{https://archive.ics.uci.edu/dataset/222/bank+marketing}: The data is related with direct marketing campaigns of a Portuguese banking institution. The marketing campaigns were based on phone calls. Often, more than one contact to the same client was required, in order to access if the product (bank term deposit) would be ('yes') or not ('no') subscribed. 

\textbf{Car} \footnote{https://archive.ics.uci.edu/dataset/19/car+evaluation}: Car Evaluation dataset derived from simple hierarchical decision model, is useful for testing constructive induction and structure discovery methods. Target is to evaluate the car : Unacceptable, Acceptable, Good or Very Good.

Apart from that, most synthesizer methods mentioned in the Section \ref{sec:synthesizers} use these datasets. This is the reason behind using the first 4 datasets for our experiments.
The next 4 datasets are popular UCI datasets with an inherent imbalance whereas the first 4 datasets are imbalanced by sampling.

\subsection{Preprocessing} 
In case of datasets with categorical columns that are not of datatype \textit{int} or \textit{float}, the categorical columns need to first be label encoded to get columns of dtype \textit{int}. \sysname method is applied on this encoded dataset to obtain the \sysname column. Once the \sysname column is obtained, one can replace the target column from the original (non-processed) dataset with the \sysname column to get the \sysname dataset. Then depending upon which synthesizer is being used, the pre-prosessing of the dataset will be done as specified for that model/ synthesizer.

\section{Synthesizers} \label{sec:synthesizers}

Among the many different methods of generation of Tabular data, there are some broad class categories: SMOTE and its variants, GAN methods, Diffusion methods, Flow based methods and LLM and Transformer methods. The current state of the art are diffusion and GAN methods closely followed by LLM based methods. 

We demonstrated the effectiveness of \sysname on five deep data generators, namely - CTGAN, CTabGAN+, TabDDPM, TabSyn and ForestFlow in our paper. We used the official GitHub implementation of each of these synthesizers. All of these generator are current/ past SOTA models and we provide a short description of each of them as follows:\\

\textbf{CTGAN}\footnote{https://github.com/sdv-dev/CTGAN}: This model introduced mode specific normalization and conditional vectors for training and 
sampling. CTGAN consists of a conditional GAN. During training, it randomly samples a discrete column based on its log frequency and learns the distribution conditioned on a label of the sampled column. The model penalizes the cross-entropy loss between the true label and the generated label.

\textbf{CTabGAN+}\footnote{https://github.com/Team-TUD/CTAB-GAN-Plus}:It is an improvement over CTabGAN which itself is an improvement over CTGAN. It consists of  extra preprocessing for mixed data types, long tailed distributions and contains an additional classifier component in the loss to learn better distributions.

\textbf{TabDDPM}\footnote{https://github.com/yandex-research/tab-ddpm}: It is a diffusion based synthetic data generation model. As all diffusion models, TabDDPM too consists of a forward and reverse diffusion process. Herein, the numerical columns are modeled by the usual Gaussian distribution process while the the categorical columns are modeled by a Multinomial diffusion process. Each column has a separate forward diffusion process while the reverse diffusion for all columns takes place together via an MLP (Multi Layer Perceptron). For each of the categorical columns, which are one hot encoded before passing through the forward diffusion process, the output of the MLP is passed through a Softmax activation which produces a probability distribution over all the classes in a column

\textbf{TabSyn}\footnote{https://github.com/amazon-science/tabsyn}: This model consists of a diffusion process modeled in a learnt latent space. The tabular data is first passed through a Variational auto-encoder which consists of a tokenizer that generates unique representations for each column which are then fed to a transformer to capture the intricate relations between the columns to obtain data in a continuous latent space instead of separately for numerical and categorical columns. The VAE is trained with a loss function involving a reconstruction term between the input and generated data and a KL divergence term which regularizes the mean and variance of the latent space.  The data in the latent space is then learn via a diffusion process and the data obtained from the reverse diffusion process is modeled back to the original space using the decoder part of the VAE.

\textbf{ForestFlow}\footnote{https://github.com/SamsungSAILMontreal/ForestDiffusion}: It is a diffusion-based model which instead of using an MLP to model the reverse diffusion process as is normally done, uses a ensemble of XGBoost models, a separate one for each discrete column and each noise level.

\subsection{Smote and its variants}
SMOTE ~\cite{Chawla_2002} was the first attempt in handling unbalanced data. It considered minority points and interpolated the minority points with its same class neighbors. There are variations of SMOTE namely BorderlineSMOTE~\cite{han2005borderline}, ADASYN~\cite{he2008adasyn} etc. which build upon the interpolation concept considering the neighbors of the minority class points. There are Kernel SMOTE for SVM classifiers \cite{ksmotesvm} and K Means SMOTE \cite{kmeansmote} which also uses clustering before interpolating like the other SMOTE Methods.

\subsection{Generative Adversarial Networks}
GANs gained increased popularity in the image generation domain. GAN (Generative Adversarial Network) has a Generator and Discriminator which work as adversaries to each other. The generator objective is to generate indistinguishable images as real data and the discriminator’s objective is to identify the real and synthetic data. The tabular data generation posed certain challenges compared to image generation. The data type of columns varied from categorical to numerical requiring different preprocessing and handling steps. The continuous data columns were not simple gaussian and could not be efficiently represented by a minmax normalization and required gaussian mixture models suited for multimodal data. With these GMMs, the problem of mode collapse becomes prominent. The first breakthrough in Tabular data generation through the GAN model was through CTGAN. \cite{ctgan} This model introduced mode specific normalization and conditional vectors for training in sampling. This method handled imbalanced data well compared to SMOTE and other interpolation methods. 
The CTABGAN was an improvement over CTGAN which added extra preprocessing for mixed data types, long tailed distributions and an additional classifier component to aid better performance \cite{ctabgan}. The CTABGAN+ is yet another improvement on CTABGAN which is currently the best method for Tabular Synthetic Data generation from imbalanced data.\cite{zhao2022ctabgan+}
Some techniques used in GAN based methods are summarized as follows:
\subsubsection{Mode Specific Normalization}
Most continuous valued attributes are multimodal in a tabular data. The mode collapse problem becomes prominent in this case and we preclude it by a method called mode specific normalization. The method includes

\begin{itemize}
    \item Construct a Variation Gaussian Mixture model and find $m$ modes in the column.
    \item For every point, find the mode to which it belongs with maximum probability and find the one hot encoded beta vector of modes.
    \item Normalize the point at that mode to get a scalar $\alpha$ with the formula:
    \[
    \alpha = \frac{{x - \mu}}{{4\sigma}}
    \]
    where $x$ is the point, $\mu$ is the mean of the mode, and $\sigma$ is the standard deviation of the mode.
    \item Concatenate for each point in a continuous attribute the beta vector and alpha scalar to obtain the embedding passed to the Neural Network stage.
\end{itemize}

\subsubsection{Conditional generator and training by sampling}

A conditional mask vector \cite{ctgan} is constructed and filled with all zeros initially and set to 1 at exactly one spot. A random categorical attribute is chosen and a value from that column is sampled from a probabilistic mass function representing the logarithmic frequency of the values in that column. This ensures that unbalanced columns get a good representation of the minority class too.

A loss is applied between the mask function and generated sample to ensure that value is generated conditionally. Let \( d_i \) be such that \( d^{(k)}_i = 0 \) for every i and k and  \( d^{(k^*)}_{i^*} = 1 \) for the real condition in mask  \( (d_{i^*} = {k^*}) \). The loss function is the cross-entropy between \( m_i \) and \( d_i \)

\subsubsection{Representation of Long tailed distributions}
Long-tailed distributions are compressed after a certain bound using the logarithmic function to capture the range of the column values \cite{ctabgan}. An example is that of the transaction amount spent in a credit card dataset; the mean and 95th percentile of the data for Transaction would be a small amount, but there would be some transactions spending one billion or above. Generative Adversarial Networks (GANs) which don’t apply this transformation are found not to be able to capture the maximum amount like that of a real dataset. The transformation applied is as follows: For such a variable having values with a lower bound \( l \), we replace each value \( \tau \) with a compressed value \( \tau_c \):
\[
\tau_c = 
\begin{cases}
\log(\tau) & \text{if } l > 0 \\
\log(\tau - l + \epsilon) & \text{if } l \leq 0
\end{cases}
\]
where \( \epsilon > 0 \). The log-transform allows to compress and reduce the tails of the distribution.

\subsubsection{Network structure}
Network in a conditional GAN uses \textit{gumbel} softmax activation for categorical and mode part of numerical attributes while it uses \textit{tanh} activation for scalar in a numerical attribute. \textit{PACGAN} method is used to reduce mode collapse problems. \cite{ctgan} 

\subsubsection{Performance comparison}
Amongst the compared GAN methods, the timeline improves performance. The CTGAN was surpassed by CTabGAN which in turn was beaten by metrics of CTABGAN+. The advantages of these considered methods are that pre-processing method are good for mixed and long tailed distributions and most importantly the formidable problem of \textbf{ imbalanced dataset} is considered and solved by the Conditional generator and \textit{training by sampling}. 

\subsection{Diffusion based methods}
Denoising Diffusion models are the most popular generative model in today’s literature. In TabDDPM, the columns which are numerical are denoised from a gaussian distribution while categorical columns are denoised from a multinomial diffusion method. A Multi-layer perceptron learns the noise added in the forward process and is used to generate rows of columnar data. Time embedding is present as an input to the MLP as in a traditional denoising diffusion model. 
\subsubsection{TABDDPM}
\begin{figure}[h]
    \centering
    \includegraphics[width=0.75\linewidth]{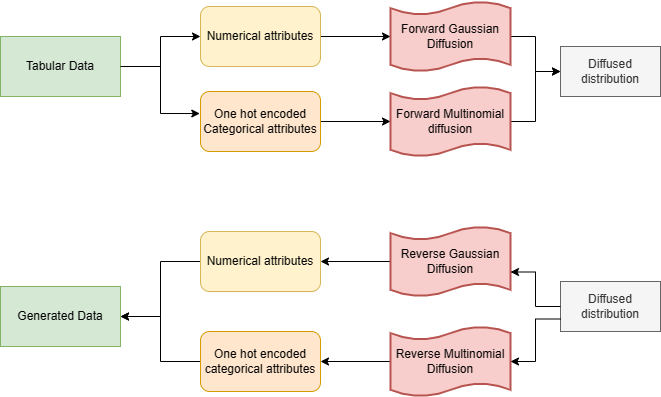}
    \caption{TABDDPM Schematic Diagram}
    \label{fig:tabddpm}
\end{figure}

A traditional application of denoising diffusion models is that of the TabDDPM. It consists of one hot encoding of categorical data which undergoes a multinomial diffusion where the numerical attributes undergo a standard gaussian denoising. \cite{tabddpm}. An idea of TABDDPM is shown in Fig \ref{fig:tabddpm}. The loss function of the diffusion is same as standard Gaussian diffusion for numerical and for discrete attributes KL divergence term as in Eqn.\ref{eq:lossddpm}.

\begin{equation} \label{eq:lossddpm}
L_{\text{TabDDPM}}(t) = L_{\text{simple}}(t) + \frac{1}{C} \sum_{i=1}^{C} \text{L}_i(t) 
\end{equation}

Where \( C \) represents the number of categorical variables.

\textbf{Result Analysis:}
TabDDPM consistently outperforms TVAE and CTABGAN+ on diverse datasets, highlighting the advantage of diffusion models for tabular data across various domains. The SMOTE method exhibits competitive performance comparable to TabDDPM and often surpasses GAN/VAE approaches, underscoring the importance of comparing against diverse methods in evaluating generative models for tabular data.
The study advocates for the use of evaluation protocols employing state-of-the-art models like CatBoost over weaker classifiers/regressors to provide more informative insights into model performance and avoid potential misinterpretation regarding the value of synthetic data generated by generative models.
\subsubsection{TABSYN}
TABSYN is a Latent diffusion model which encodes the data in a latent space and performs diffusion in the latent space. The claim is that the correlation between columns is better preserved in the latent space than diffusion in TabDDPM. It has a VAE to encode data to latent space and a diffusion model which denoises. It consists of a transformer based Beta-VAE which biases reconstruction error loss component over the KL divergence component. The embedding input to the transformer is done by a tokenizer which applies a linear embedding to numerical and one hot embedding lookup for categorical data attributes. Diffusion then takes place in the latent dimension and since diffusion is resistant to noise in the data, we can apply differential privacy based methods in the encoder part of the network. Loss Function of the Autoencoder is given by Eqn.\ref{losstabsyn}.

\begin{equation}\label{losstabsyn}
  L = \ell_{\text{recon}}(x, \hat{x}) + \beta \ell_{\text{kl}} 
  \end{equation}
  
where $\ell_{\text{recon}}$ is the reconstruction loss between the input data and the reconstructed one, and $\ell_{\text{kl}}$ is the KL divergence loss that regularizes the mean and variance of the latent space.


\begin{figure}[h]
\centering
    \includegraphics[width=0.75\linewidth]{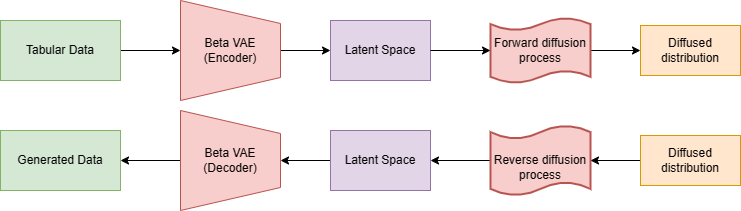}
    \caption{TABSYN Schematic Diagram}
    \label{fig:TABSYN}
\end{figure}
\textbf{Result analysis}
The experiments compared TABSYN with seven existing tabular data generation methods, including classical GAN and VAE models, and recent state-of-the-art methods. TABSYN consistently outperformed all baseline methods in a standardized setting, demonstrating its superiority across various tasks.

\subsection{Transformer Based Methods}
Transformer based methods and LLMs have been occupying high prominence accomplishing incredibly hard NLP tasks reasonably. GReaT (Generation of Realistic Tabular data) model encodes numerical and categorical data. The conditioning of columns is done with a subset of columns already available. Then it permutes the columns in random order for training all possible conditioning of the columnar data. This conditioning is an advantage of this method over GAN and diffusion-based methods. TabuLA model speaks of training a foundational model for tabular data instead of using pre-trained GPT weights. This model enhances the quality of tabular data generated as and when more tabular datasets are trained with this model.
Advantages claimed by these methods are incorporation of pre trained context knowledge in better learning dependencies and joint distributions. Another important advantage is that of simplicity and easily Interpretability  of direct output. One prominent disadvantage is scaling for large datasets where token limit exceeds. 

\subsubsection{GReaT} 
GReaT (Generation of Realistic Tabular data) \cite{GREAT} is a model which uses the state-of-the-art GPT for its conditional generation of tabular data. This model encodes numerical and categorical data together. The conditioning of  columns is done with a subset of columns already available. Then it permutes the columns in random order for training all possible conditioning of the columnar data. This type of conditioning is its claim over GAN based methods. 

Pretrained GPT2/ DistilGPT2 model is finetuned for our required dataset. The training data is encoded as a sentence which can be understood by a LLM. The sentence structure is a simple “Col1 is Val1, Col2 is Val2 …”. Random order of columns is passed in training to learn all possible conditional distributions of different columns given a fixed subset of other columns.

\subsubsection{TabuLa} 
\begin{figure}[h]
    \centering
    \includegraphics[width=0.5\linewidth]{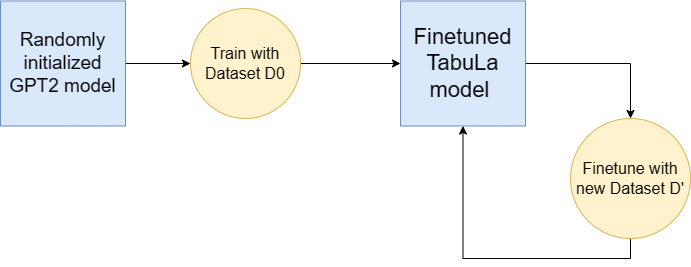}
    \caption{TabuLa - Schematic figure showing repeated finetuning with new tabular datasets and finally obtaining a foundational model}
    \label{fig:tabula}
\end{figure}

This speaks of training a foundational model for tabular data instead of going with pre-trained GPT weights \cite{tabula}. This has the advantage that tokens can be highly optimized compared to using a pre trained LLM. The analysis has found that the more the datasets on which model is finetuned, the better the performance and utility of the synthetic data generated. The method claimed to be better than the then diffusion and GAN models. The schematic diagram of TabuLa is shown in Fig \ref{fig:tabula}

\subsubsection{TABMT}
Tabular Masked transformers (TABMT) \cite{gulati2023tabmt} can be trained to predict column values in each row of a tabular dataset. This model has a good balance between privacy and quality. Random order of masks prevent bias caused by position in columnar data. Masking probability from a uniform distribution causes a masked set of all sizes to arise during training. This makes the training and generation very similar. The generation is by sampling column by column which might be a little slower than other methods. Architecture consists of a transformer encoder which takes in embeddings of each column and a dynamic Linear layer at the end which can output logits of different tokens in each column. The preprocessing involves one-hot encoding the categorical values whereas the numerical values are lossily segregated by clustering. 
\textbf{Result Analysis:} For 15 datasets, TABMT performs better than Baselines CTABGAN+, TABDDPM, RealTabFormer and TVAE. 

\subsubsection{Performance Comparison}
Amongst all the Transformer based synthesizers discussed so far, best performance is from TABMT. The next in line is the TabuLa and finally the base GReaT method. Overall, the conditional generation is the claim of LLM and Transformer based methods. Privacy metric through Distance to the Closest Record is also shown to be better than their contemporaneous models. \cite{gulati2023tabmt} \cite{GREAT}. Since code for TABMT is not available, our implementation of TABMT did not exceed the performance of GAN and diffusion models and hence we did not use TABMT as a baseline.

\subsection{Common Metrics of Evaluation}

\subsubsection{MLE - Machine Learning Efficiency}
It quantifies the performance of classification or regression models \textit{trained on synthetic data and evaluated on the real test set}. The model developed using the augmented data should exhibit good accuracy, close to the original model accuracy (or even better in the case of imbalanced data). This demonstrates the utility of the generated data.

\subsubsection{Distance to Closest Record Histogram}
This measure ensures that while synthetic data should resemble real data, it should not be an exact copy. The distance to the closest record in the training data (real) from the synthetic data is calculated. Ideally, this distribution should resemble that between test data (real) and training data (real).

\subsubsection{Statistical Similarity}
The correlation between columns of the original and synthetic data should be similar. The marginal probabilities of individual columns should be captured while also fulfilling the joint distribution to be matched.

The motivation for using Machine Learning Efficacy among other metrics is because of the problem statement involving the goal of increasing the accuracy of a trained classifier in an imbalanced setting. Hence, we only consider the metric Machine Learning Efficacy i.e. measuring the classifier accuracy when model is trained on synthetic data.

%% file: main.bbl
\begin{thebibliography}{28}
\providecommand{\natexlab}[1]{#1}

\bibitem[{Adiputra and Wanchai(2023)}]{CTGANENN}
Adiputra, I. N.~M.; and Wanchai, P. 2023.
\newblock CTGAN-ENN: A tabular GAN-based Hybrid Sampling Method for Imbalanced and Overlapped Data in Customer Churn Prediction.

\bibitem[{Borisov et~al.(2023)Borisov, Sessler, Leemann, Pawelczyk, and Kasneci}]{GREAT}
Borisov, V.; Sessler, K.; Leemann, T.; Pawelczyk, M.; and Kasneci, G. 2023.
\newblock Language Models are Realistic Tabular Data Generators.
\newblock In \emph{The Eleventh International Conference on Learning Representations}.

\bibitem[{Breiman(2001)}]{randomforest}
Breiman, L. 2001.
\newblock Random forests.
\newblock \emph{Machine learning}, 45: 5--32.

\bibitem[{Chawla et~al.(2002)Chawla, Bowyer, Hall, and Kegelmeyer}]{Chawla_2002}
Chawla, N.~V.; Bowyer, K.~W.; Hall, L.~O.; and Kegelmeyer, W.~P. 2002.
\newblock SMOTE: Synthetic Minority Over-sampling Technique.
\newblock \emph{Journal of Artificial Intelligence Research}, 16: 321–357.

\bibitem[{Chen and Guestrin(2016)}]{chen2016xgboost}
Chen, T.; and Guestrin, C. 2016.
\newblock Xgboost: A scalable tree boosting system.
\newblock In \emph{Proceedings of the 22nd acm sigkdd international conference on knowledge discovery and data mining}, 785--794.

\bibitem[{Douzas, Bacao, and Last(2018)}]{kmeansmote}
Douzas, G.; Bacao, F.; and Last, F. 2018.
\newblock Improving imbalanced learning through a heuristic oversampling method based on k-means and SMOTE.
\newblock \emph{Information Sciences}, 465: 1--20.

\bibitem[{Gulati and Roysdon(2023)}]{gulati2023tabmt}
Gulati, M.~S.; and Roysdon, P.~F. 2023.
\newblock TabMT: Generating tabular data with masked transformers.
\newblock arXiv:2312.06089.

\bibitem[{Han, Wang, and Mao(2005)}]{han2005borderline}
Han, H.; Wang, W.-Y.; and Mao, B.-H. 2005.
\newblock Borderline-SMOTE: a new over-sampling method in imbalanced data sets learning.
\newblock In \emph{International conference on intelligent computing}, 878--887. Springer.

\bibitem[{He et~al.(2008)He, Bai, Garcia, and Li}]{he2008adasyn}
He, H.; Bai, Y.; Garcia, E.~A.; and Li, S. 2008.
\newblock ADASYN: Adaptive synthetic sampling approach for imbalanced learning.
\newblock In \emph{2008 IEEE international joint conference on neural networks (IEEE world congress on computational intelligence)}, 1322--1328. Ieee.

\bibitem[{Hegselmann et~al.(2023)Hegselmann, Buendia, Lang, Agrawal, Jiang, and Sontag}]{tabllm}
Hegselmann, S.; Buendia, A.; Lang, H.; Agrawal, M.; Jiang, X.; and Sontag, D. 2023.
\newblock Tabllm: Few-shot classification of tabular data with large language models.
\newblock In \emph{International Conference on Artificial Intelligence and Statistics}, 5549--5581. PMLR.

\bibitem[{Johnson and Khoshgoftaar(2019)}]{dataGenMethods}
Johnson, J.~M.; and Khoshgoftaar, T.~M. 2019.
\newblock Survey on deep learning with class imbalance.
\newblock \emph{Journal of big data}, 6(1): 1--54.

\bibitem[{Jolicoeur-Martineau, Fatras, and Kachman(2024)}]{forestflow}
Jolicoeur-Martineau, A.; Fatras, K.; and Kachman, T. 2024.
\newblock Generating and imputing tabular data via diffusion and flow-based gradient-boosted trees.
\newblock In \emph{International Conference on Artificial Intelligence and Statistics}, 1288--1296. PMLR.

\bibitem[{Kotelnikov et~al.(2022)Kotelnikov, Baranchuk, Rubachev, and Babenko}]{tabddpm}
Kotelnikov, A.; Baranchuk, D.; Rubachev, I.; and Babenko, A. 2022.
\newblock TabDDPM: Modelling Tabular Data with Diffusion Models.
\newblock arXiv:2209.15421.

\bibitem[{Leng et~al.(2024)Leng, Guo, Tao, Meng, and Wang}]{OBMI}
Leng, Q.; Guo, J.; Tao, J.; Meng, X.; and Wang, C. 2024.
\newblock OBMI: oversampling borderline minority instances by a two-stage Tomek link-finding procedure for class imbalance problem.
\newblock \emph{Complex \& Intelligent Systems}, 1--18.

\bibitem[{Li et~al.(2021)Li, Huang, Liu, and Jiang}]{li2021hybrid}
Li, Z.; Huang, M.; Liu, G.; and Jiang, C. 2021.
\newblock A hybrid method with dynamic weighted entropy for handling the problem of class imbalance with overlap in credit card fraud detection.
\newblock \emph{Expert Systems with Applications}, 175: 114750.

\bibitem[{Mathew et~al.(2015)Mathew, Luo, Pang, and Chan}]{ksmotesvm}
Mathew, J.; Luo, M.; Pang, C.~K.; and Chan, H.~L. 2015.
\newblock Kernel-based SMOTE for SVM classification of imbalanced datasets.
\newblock In \emph{IECON 2015 - 41st Annual Conference of the IEEE Industrial Electronics Society}, 001127--001132.

\bibitem[{Papamakarios, Pavlakou, and Murray(2017)}]{MAF}
Papamakarios, G.; Pavlakou, T.; and Murray, I. 2017.
\newblock Masked autoregressive flow for density estimation.
\newblock \emph{Advances in neural information processing systems}, 30.

\bibitem[{Vaswani(2017)}]{transformer}
Vaswani, A. 2017.
\newblock Attention is all you need.
\newblock \emph{arXiv preprint arXiv:1706.03762}.

\bibitem[{Vuttipittayamongkol and Elyan(2020)}]{vuttipittayamongkol2020neighbourhood}
Vuttipittayamongkol, P.; and Elyan, E. 2020.
\newblock Neighbourhood-based undersampling approach for handling imbalanced and overlapped data.
\newblock \emph{Information Sciences}, 509: 47--70.

\bibitem[{Wang et~al.(2019)Wang, Wu, Zheng, Niu, and Wang}]{SMOTEtomek}
Wang, Z.; Wu, C.; Zheng, K.; Niu, X.; and Wang, X. 2019.
\newblock SMOTETomek-based resampling for personality recognition.
\newblock \emph{IEEE access}, 7: 129678--129689.

\bibitem[{Xu et~al.(2019)Xu, Skoularidou, Cuesta-Infante, and Veeramachaneni}]{ctgan}
Xu, L.; Skoularidou, M.; Cuesta-Infante, A.; and Veeramachaneni, K. 2019.
\newblock Modeling Tabular data using Conditional GAN.
\newblock In \emph{Neural Information Processing Systems}.

\bibitem[{Zhang et~al.(2024)Zhang, Zhang, Srinivasan, Shen, Qin, Faloutsos, Rangwala, and Karypis}]{zhang2024mixedtype}
Zhang, H.; Zhang, J.; Srinivasan, B.; Shen, Z.; Qin, X.; Faloutsos, C.; Rangwala, H.; and Karypis, G. 2024.
\newblock Mixed-Type Tabular Data Synthesis with Score-based Diffusion in Latent Space.
\newblock arXiv:2310.09656.

\bibitem[{Zhang, Zhang, and Wang(2010)}]{zhang2010cluster}
Zhang, Y.-P.; Zhang, L.-N.; and Wang, Y.-C. 2010.
\newblock Cluster-based majority under-sampling approaches for class imbalance learning.
\newblock In \emph{2010 2nd IEEE International Conference on Information and Financial Engineering}, 400--404. IEEE.

\bibitem[{Zhao et~al.(2024)Zhao, Guan, Xue, and Pan}]{HSCGS}
Zhao, X.; Guan, S.; Xue, Y.; and Pan, H. 2024.
\newblock HS-CGK: A Hybrid Sampling Method for Imbalance Data Based on Conditional Tabular Generative Adversarial Network and K-Nearest Neighbor Algorithm.
\newblock \emph{Computing and Informatics}, 43(1): 213--239.

\bibitem[{Zhao, Birke, and Chen(2023)}]{tabula}
Zhao, Z.; Birke, R.; and Chen, L. 2023.
\newblock TabuLa: Harnessing Language Models for Tabular Data Synthesis.
\newblock arXiv:2310.12746.

\bibitem[{Zhao et~al.(2022)Zhao, Kunar, Birke, and Chen}]{zhao2022ctabgan+}
Zhao, Z.; Kunar, A.; Birke, R.; and Chen, L.~Y. 2022.
\newblock CTAB-GAN+: Enhancing Tabular Data Synthesis.
\newblock arXiv:2204.00401.

\bibitem[{Zhao et~al.(2021)Zhao, Kunar, der Scheer, Birke, and Chen}]{ctabgan}
Zhao, Z.; Kunar, A.; der Scheer, H.~V.; Birke, R.; and Chen, L.~Y. 2021.
\newblock {CTAB-GAN:} Effective Table Data Synthesizing.
\newblock \emph{CoRR}, abs/2102.08369.

\bibitem[{Zhou and Liu(2005)}]{costsensitive}
Zhou, Z.-H.; and Liu, X.-Y. 2005.
\newblock Training cost-sensitive neural networks with methods addressing the class imbalance problem.
\newblock \emph{IEEE Transactions on knowledge and data engineering}, 18(1): 63--77.

\end{thebibliography}
